%% file: neurips_2026.tex
\documentclass{article}

    \PassOptionsToPackage{numbers, compress}{natbib}

\usepackage[main, final]{neurips_2026}

\usepackage[utf8]{inputenc} 
\usepackage[T1]{fontenc}    
\usepackage{hyperref}       
\usepackage{url}            
\usepackage{booktabs}       
\usepackage{amsfonts}       
\usepackage{nicefrac}       
\usepackage{microtype}      

\usepackage{graphicx}
\usepackage{multirow}
\usepackage{amsmath}

\usepackage[table]{xcolor}
\usepackage{float}      
\usepackage{makecell}   
\usepackage{soul}       
\usepackage{enumitem}   
\usepackage{wrapfig}
\usepackage{algorithm}
\usepackage{algorithmic}
\usepackage{caption}

\usepackage{subcaption} 

\title{Concepts in Motion: Temporal Concept Bottleneck Model for Interpretable Video Classification}

\author{%
\begin{tabular}{c}
\bfseries
Patrick Knab\textsuperscript{1,2},
Sascha Marton\textsuperscript{1},
Philipp J. Schubert\textsuperscript{2}, \\
\bfseries
Drago Guggiana\textsuperscript{2},
Christian Bartelt\textsuperscript{1}
\\[1ex]
\normalfont\textsuperscript{1}Clausthal University of Technology\quad
\normalfont\textsuperscript{2}Ramblr.ai Research \\
\texttt{patrick.knab@tu-clausthal.de}
\end{tabular}
}

\begin{document}
\addtocontents{toc}{\protect\setcounter{tocdepth}{-1}}

\maketitle

\begin{abstract}
Concept Bottleneck Models (CBMs) enable interpretable image classification by structuring predictions around human-understandable concepts, but extending this paradigm to video remains challenging due to the difficulty of extracting concepts and modeling them over time.
In this paper, we introduce \textbf{MoTIF} (Moving Temporal Interpretable Framework), a transformer-based concept architecture that operates on sequences of temporally grounded concept activations, by employing per-concept temporal self-attention to model when individual concepts recur and how their temporal patterns contribute to predictions.
Central to the framework is a class-conditioned VLM-based concept discovery module that extracts object- and action-centric textual concepts from training videos, yielding temporally expressive concept sets without manual concept annotation.
Across multiple video benchmarks, this combination improves over global concept bottlenecks and remains competitive within the interpretable concept-bottleneck setting, while narrowing the gap to strong black-box video baselines that we report as contextual references.
\noindent Our code is available at \url{https://github.com/patrick-knab/MoTIF}

\end{abstract}

\input{sections/intro}
\input{sections/motif}
\input{sections/expos}
\input{sections/related}
\input{sections/conclusion}

\section*{Acknowledgments}
This research was supported in part by the German Federal Ministry for Economic Affairs and Climate Action (BMWK) and the German Federal Ministry for Research, Technology and Space (BMFTR). Part of the work was conducted during an internship at Ramblr.ai.

\textbf{Competing interests.} Patrick Knab was an intern at Ramblr.ai during this work. The authors declare no other competing interests.

\bibliographystyle{unsrtnat}
\bibliography{example_paper}


\appendix

\newpage
\addtocontents{toc}{\protect\setcounter{tocdepth}{2}}
\renewcommand{\contentsname}{Appendix Contents}
\tableofcontents

\input{sections/appendix}


\newpage
\input{sections/checklist}

\end{document}

%% file: sections/intro.tex
\section{Introduction}
Modern deep learning models already achieve outstanding results in video understanding tasks such as video classification, action recognition, and event detection \citep{noframe, timesformer}.
Despite their success, these models are commonly perceived as \textit{black boxes} since their internal workings are not interpretable in a way that reveals their decision-making process \citep{molnar, lime_survey}.
Concept Bottleneck Models (CBMs) \citep{cbm_original} address this issue by enforcing an intermediate bottleneck layer of human-understandable concepts, which are then used by a linear classifier to generate the final prediction.

\begin{figure}[h]
  \centering
  \includegraphics[width=\linewidth]{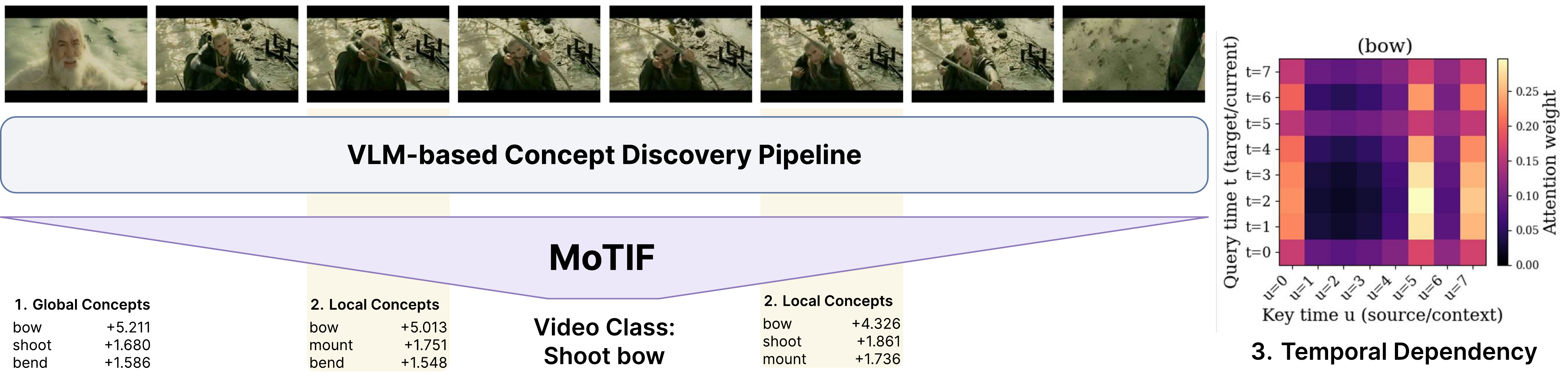}
    \caption{\textbf{MoTIF.} The method treats videos as windows and produces local concept explanations, global explanations for entire videos, and temporal dependency maps from the attention heads. The model is MoTIF (ViT-L/14) and sample frames are from HMDB51 \citep{hmdb}, licensed under CC BY 4.0.}
  \label{fig:start}
\end{figure}

While CBMs have been extensively studied in the image domain \citep{dcbm, labo, llmcbm, schrodi}, their extension to video remains largely unexplored \citep{jeyakumar2022automatic}.
Videos differ from images in that they contain a \textit{temporal component}: concepts evolve over time, and many actions cannot be inferred from a single frame \citep{pcbear, vlwm}.
Importantly, the challenge of variable video length is orthogonal to modeling long-range dependencies.
While transformers \citep{attention} excel at capturing such dependencies \citep{timesformer}, their dense temporal feature mixing obscures concept-level attributions, limiting their suitability for interpretable concept bottleneck modeling \citep{molnar, hao2021self}. This highlights the need for architectures that handle variable-length inputs while preserving concept-level interpretability over time.

In this work, we introduce \textbf{MoTIF} (\textbf{Mo}ving \textbf{T}emporal \textbf{I}nterpretable \textbf{F}ramework), a concept bottleneck model tailored for video classification.
MoTIF builds on transformer-inspired blocks and introduces a \textit{per-channel temporal self-attention} (diagonal attention) mechanism that isolates temporal reasoning for each concept to enable interpretation.
In addition, we utilize a VLM-based concept discovery approach to generate object- and action-based, class-conditioned textual concepts from training videos.
To illustrate how MoTIF extends beyond static images, Figure~\ref{fig:start} shows our framework: it processes a video, tracks its concepts through time, and explains which concepts drive the final prediction.
MoTIF tracks concepts such as \emph{bow}, \emph{mount}, and \emph{shoot} over time, whose combined activations support the final prediction.
Temporal dependency maps indicate which windows provide contextual support for updating a concept channel, and are interpreted jointly with concept presence and contribution rather than as standalone evidence.
This yields both \textit{global concept contributions} over the full video and \textit{local relevance} within specific windows.

Our key contributions are:
\begin{itemize}
  \item \emph{MoTIF}, a Video CBM with per-channel temporal self-attention that \textbf{preserves concept independence} within Transformer blocks while modeling temporal dynamics at the level of individual concepts.
  \item \emph{MoTIF} provides \textbf{three complementary explanation modes}: \textit{(i)} global concept relevance via log-sum-exp (LSE) pooling, \textit{(ii)} localized temporal explanations using windowed concept attributions, and \textit{(iii)} attention-based temporal maps that visualize how a concept channel distributes its focus across time.
  \item A \textbf{VLM-based concept discovery pipeline} to extract textual object and action concepts directly from training videos with weak supervision from class labels, improving concept coverage and downstream performance over global bottlenecks.
\end{itemize}

%% file: sections/motif.tex
\section{MoTIF}
\label{sec:method}

\begin{figure}[t]
    \centering
    \includegraphics[width=\linewidth]{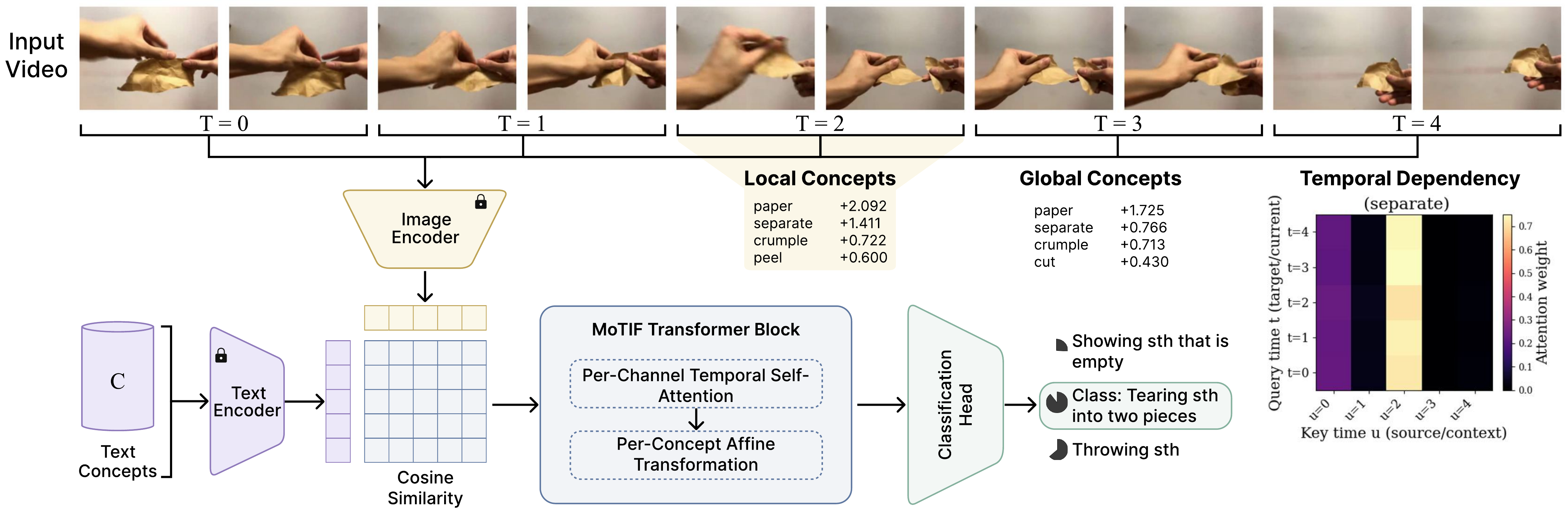}
    \vspace{0.5em}
    \caption{\textbf{MoTIF pipeline.}
    Videos are embedded with a vision--language backbone and mapped to concept activations via cosine similarity.
    Per-channel temporal self-attention models dynamics independently for each concept, followed by a per-concept affine transformation, and classification.
    MoTIF enables explanations across three views: global concepts, local concepts, and temporal dependencies. Sample frames from SSv2 \citep{ssv2} with MoTIF (ViT-L/14).}
    \label{fig:method-overview}
\end{figure}

\textbf{MoTIF} is a transformer-based CBM for video classification that operates on sequences of window-level concept activations (Figure~\ref{fig:method-overview}).
A video is split into $T$ temporal windows $w_t$, each embedded by a frozen vision backbone, and compared against a concept bank $\mathcal{C}$ of VLM-generated object and action concepts.
The resulting cosine similarities form the activation matrix $X \in \mathbb{R}^{T \times C}$, where $C = |\mathcal{C}|$, which is processed by the temporal module before LSE pooling and classification.

The framework supports concept-wise attribution through concept-specific temporal processing: temporal dependencies are modeled independently for each concept channel, so recurrent temporal patterns---\emph{motifs}---remain attributable to individual concepts.
We use the term \emph{channel} to denote one dimension of the concept space.
Each video $n$ yields a class label $y^{(n)} \in \{1,\ldots,K\}$ and concept activations $X^{(n)} \in \mathbb{R}^{T_n \times C}$, where the second axis ($C$) indexes concepts; each slice $X^{(n)}_{:,c} \in \mathbb{R}^{T_n}$ corresponds to the temporal activation sequence of concept $c \in \{1,\ldots,C\}$.
Unlike CNN channels, these dimensions are semantically interpretable by design.
After temporal processing, per-concept activations are refined via a per-concept affine transformation followed by a Softplus nonlinearity, and a classifier produces per-time-step logits.
Video-level predictions are obtained via log-sum-exp pooling, which also yields a time-importance profile for additional explanation.
In contrast, a conventional transformer mixes channels during attention \citep{attention}\footnote{We also evaluate a variant with \emph{full multi-head attention}, but emphasize that this removes explicit concept attribution.
A detailed analysis of this and other design choices is provided in Section~\ref{sec:ablation} and Appendix~\ref{full_attention}.}.
Algorithms are given in Appendix~\ref{app:algorithms}.

\subsection{VLM-based concept discovery}
\label{sec:agentic_concepts}

\textbf{Concept discovery.}
The concept discovery stage constructs a textual concept bank from training videos, which is then used to compute the structured, interpretable activations fed to the MoTIF backbone.
While formulated for video, images are naturally handled as the special case of a single temporal window.
In our main setting, we sample a small number of training videos per class and process each window $w_t$ with a VLM $A_{\mathrm{vlm}}$ conditioned on the class label, yielding natural-language concept candidates
\begin{equation}
(\mathcal{C}^{\text{obj}}_{w_t}, \mathcal{C}^{\text{act}}_{w_t})
= A_{\mathrm{vlm}}(w_t, y),
\end{equation}
capturing objects and actions.
The discovery VLM is used only at this offline stage: test videos are not passed to it at inference, but are mapped to the fixed concept bank using the embedding backbone.
This class-conditioned procedure requires no manually specified concept vocabulary, and the discovery model can be replaced to suit different tasks.

\textbf{Concept and video embeddings.}
The global concept bank $\mathcal{C}$ is constructed from the union of all discovered textual object and action concepts.
To ensure semantic diversity and remove near-duplicates, concepts are embedded using the text encoder of the CLIP-based backbone \citep{clip} that MoTIF builds on, and filtered by cosine similarity: concepts with pairwise similarity greater than $0.9$ are removed.
For each window, depending on the backbone, visual embeddings are obtained either from a representative frame or a video-adapted CLIP variant (Appendix~\ref{back:integration}).
Concept activations are computed as cosine similarities between window embeddings and concept embeddings, yielding per-video activation matrices
$X^{(n)} \in \mathbb{R}^{T_n \times C}$,
where each channel corresponds to one interpretable concept and is processed independently.
Appendix~\ref{app:concepts} provides evidence that concept quality matters: across discovery backbones, filtering thresholds, concept sources, and object/action splits, downstream performance changes systematically with the quality and diversity of the concept bank.

\subsection{MoTIF's transformer bottleneck}

\subsubsection{Per-channel temporal self-attention}
In standard transformers, query--key--value (QKV) projections are implemented as full linear layers ($W_Q \in \mathbb{R}^{C \times C}$), which mix channels and would obscure concept attribution.
In MoTIF, we avoid mixing: before applying diagonal temporal attention, we add a fixed sinusoidal positional encoding to the concept activation sequence to encode temporal order. Each concept $c$ then receives its own QKV projections via depthwise $1{\times}1$ convolutions,

\begin{equation}
Q, K, V \in \mathbb{R}^{T \times C},
Q_{:,c} = X_{:,c}\,\theta_Q^{(c)},
K_{:,c} = X_{:,c}\,\theta_K^{(c)},
V_{:,c} = X_{:,c}\,\theta_V^{(c)}.
\end{equation}

Here \(\theta_{Q}^{(c)},\theta_{K}^{(c)},\theta_{V}^{(c)} \in\mathbb{R}\) are channel-specific scalars applied uniformly across all \(T\), not temporal filters.
Equivalently, each projection uses a depthwise kernel \(\Theta_Q,\Theta_K,\Theta_V\in\mathbb{R}^{C\times1\times1}\), where \(\theta_{*}^{(c)}\) is the \(c\)-th depthwise filter.
Attention scores are computed \emph{per concept} as $W_{c,t,u} \;=\; Q_{t,c}\,K_{u,c}$, where $t$ denotes the query time step and $u$ the key/value time step.
A softmax over $u$ yields attention weights $W \in \mathbb{R}^{C \times T \times T}$, so that each concept decides \emph{which of its past or future activations to attend to}.
The output is obtained as the weighted sum over $V$:
$X^{(L)}_{t,c} = \sum_{u=1}^T W_{c,t,u} \, V_{u,c}$.
Here, $L$ denotes the number of temporal transformer layers; the main configuration uses $L=1$.
This diagonal attention structure supports concept attribution because the attention operation itself does not directly mix concept channels.

\begin{equation}
\setlength{\arraycolsep}{3pt}
\begin{array}{cc}
\underbrace{
\left[
\begin{smallmatrix}
(c_1 \to\  c_1) & \cdots & (c_1 \to\ c_C)\\
\vdots            & \ddots & \vdots \\
(c_C \to\  c_1) & \cdots & (c_C \to\  c_C)
\end{smallmatrix}
\right]
}_{\text{Full attention}}
&
\underbrace{
\left[
\begin{smallmatrix}
(c_1 \to\  c_1) &        &        \\
                  & \ddots &        \\
                  &        & (c_C \to\  c_C)
\end{smallmatrix}
\right]
}_{\text{Diagonal attention}}
\end{array}
\label{eq:4}
\end{equation}

The block concludes with LayerNorm across concepts at each time step.
In this work, we deliberately keep the learned temporal transformations per concept to support clear concept-level attribution, while shared normalization statistics introduce limited cross-concept coupling.
Importantly, this is a design choice rather than a hard restriction: as shown in Section~\ref{sec:ablation}, enabling explicit cross-concept interactions can improve accuracy while making individual concept attributions less direct.

\textbf{Architectural extension.}
For most experiments in this paper, we report the results of MoTIF using diagonal attention within a standard transformer architecture.
However, as shown by \cite{timesformer}, separating spatial and temporal attention can further enhance performance.
Therefore, we additionally evaluate MoTIF when extended to a space-time transformer architecture (MoTIF-ST), as detailed in Appendix~\ref{spaceTime}, demonstrating that MoTIF is not restricted to a single transformer design.

\textbf{Complexity.}
Diagonal attention avoids dense channel mixing, but computes a full $T \times T$ attention map for each concept channel, yielding attention-map compute and memory complexity of $\mathcal{O}(CT^2)$.
By contrast, standard multi-head attention maintains one such map per head, resulting in attention-map memory $\mathcal{O}(HT^2)$ with $H \ll C$, while additionally incurring dense channel-mixing projection costs.
Thus, diagonal attention trades efficiency for concept-specific temporal attention when the number of concepts is large (see Appendix~\ref{resources}).

\textbf{Per-concept affine transformation.}
Each refined activation $X^{(L)}_{t,c}$ can optionally be scaled and shifted by learnable concept-specific parameters, $\tilde{X}_{t,c} = \gamma_c X^{(L)}_{t,c} + \delta_c$, and then passed through a Softplus nonlinearity $Z_{t,c} = \mathrm{Softplus}(\tilde{X}_{t,c})$, which ensures nonnegative concept activations while avoiding dead units and maintaining differentiability everywhere.
This  transformation introduces a per-concept scale ($\gamma_c$) and bias ($\delta_c$), allowing the model to adapt to differences in concept magnitude and activation thresholds.

\subsubsection{Classification head}
From these activations, per-time-step logits are computed as
$\ell_t = W_{\mathrm{cls}} Z_{t,:}^{\top} + b$ with $W_{\mathrm{cls}} \in \mathbb{R}^{K \times C}$ and $b\in\mathbb{R}^{K}$, where $K$ denotes the number of target classes and $C$ the number of concepts.
Since videos vary in length, we apply \textit{log-sum-exp (LSE) pooling} across time \citep{wang2018revisiting}, which provides a smooth temporal aggregation and serves as a soft approximation to max-pooling, becoming sharper as temperature $\tau \to 0$:
\begin{equation}
\hat{c} = \tau \log \sum_{t=1}^T m_t e^{c_t/\tau},
\qquad
\hat{\ell} = \tau \log \sum_{t=1}^T m_t e^{\ell_t/\tau},
\end{equation}
where $c_t=Z_{t,:}$ is the concept vector at window $t$, $m_t\in\{0,1\}$ masks padded windows, and the exponential and logarithm are applied elementwise to vectors.
We denote the pooled concept vector by $\hat{c}$ and the pooled logits by $\hat{\ell}$.
The pooled logits $\hat{\ell}$ form the video-level prediction.

\textbf{Training objective.}
The model is trained with class-weighted cross-entropy on $\hat{\ell}$, complemented with two regularizers:
an $\ell_1$ penalty on $W_{\mathrm{cls}}$ to encourage sparsity, and an activation sparsity penalty on $Z$:
\begin{equation}
\mathcal{L}
= \mathrm{CE}(\hat{\ell}, y)
+ \lambda_{\ell_1}\,\lVert W_{\mathrm{cls}} \rVert_1
+ \lambda_{\mathrm{sparse}}
\;\frac{1}{(\sum_t m_t)C}
\sum_{t,c} m_t \lvert Z_{t,c} \rvert .
\end{equation}

\subsubsection{Explanation generation}
\label{sec:explanation}
To make predictions transparent, MoTIF decomposes them into time- and concept-resolved contributions.
For a target class $k$, the contribution of each time step is $c_t^{(k)} = Z_{t,:}\odot W_{\mathrm{cls},k,:}$, with score $s_t^{(k)} = \sum_{c=1}^C c_{t,c}^{(k)} + b_k$.
Temporal importance weights are derived consistently with LSE pooling:
\begin{equation}
\pi_t^{(k)} = \frac{m_t\exp(s_t^{(k)}/\tau)}{\sum_{u=1}^T m_u\exp(s_u^{(k)}/\tau)}.
\end{equation}
Aggregating over time yields global concept attributions, $\bar{c}^{(k)} = \sum_{t=1}^T \pi_t^{(k)} \,c_t^{(k)}$.
MoTIF therefore provides three complementary views:
\textbf{(1) Global concepts}, via $\bar{c}^{(k)}$;
\textbf{(2) Local concepts}, active in high-weight windows (large $\pi_t^{(k)}$);
\textbf{(3) Temporal dependencies}, revealed by per-concept attention maps ($W_{c,t,u}$) that show how occurrences of concepts relate across time (see Appendix \ref{app:dependencies}).  Temporal attention peaks should not be interpreted in isolation as decisive visual evidence.
Rather, the map indicates which temporal context is used when updating a given concept channel.
We therefore interpret temporal maps jointly with (i) concept presence in the queried window, (ii) its contribution to the predicted class, and (iii) the temporal windows providing contextual support.
Together, these expose both \emph{which} concepts mattered and \emph{when} they were decisive.

%% file: sections/expos.tex
\section{Experiments}

\subsection{Experimental setup}

\textbf{Datasets.}
We evaluate on Breakfast Actions~\citep{breakfast}, HMDB51~\citep{hmdb}, UCF101~\citep{ucf101}, and Something-Something V2 (SSv2)~\citep{sth-sth, ssv2}, containing 10, 51, 101, and 174 classes, respectively. Together, they cover short and long videos, local and global temporal dependencies, and different levels of action granularity and viewpoint diversity.

\textbf{Backbones.}
We use CLIP-based visual backbones with different architectures, scales, and temporal adaptation: CLIP RN/50, ViT-B/32, and ViT-L/14~\citep{clip}, SigLIP ViT-L/14~\citep{siglip}, and the video-adapted Perception Encoder with ViT-L/14 and ViT-G/14 variants~\citep{bolya2025perceptionencoderbestvisual}.

\textbf{Concept discovery.}
For VLM-based concept discovery (Section~\ref{sec:agentic_concepts}), we use Qwen~3 30B~\citep{yang2025qwen3technicalreport} to generate the concept sets used in the main experiments.
As in prior CBMs, downstream performance depends on the quality and coverage of the resulting concept bank, especially for domain-specific and dynamic concepts.
For ablations, we follow \citet{labo} and construct smaller dataset-specific concept sets using GPT-5; Appendix~\ref{app:concepts} reports a targeted concept-bank analysis covering discovery backbones, filtering thresholds, concept types, concept-set variance, and a DCBM-style \citep{dcbm} visual-concept robustness check based on SAM3 \citep{carion2026sam} segments and short visual snippets.

\begin{wraptable}{r}{0.55\textwidth}
\vspace{-1em}
\centering
\scriptsize
\caption{\textbf{Top-1 accuracy (\%).}
Mean $\pm$ standard deviation across train-test splits. Full results, including all backbones, are in Appendix~\ref{sec:ablation_concept_set}.}
\label{tab:perf_comparison}
\resizebox{\linewidth}{!}{%
\begin{tabular}{@{}lcccc@{}}
\toprule
\textbf{Method} & \textbf{Breakfast} & \textbf{HMDB51} & \textbf{UCF101} & \textbf{SSv2} \\
\midrule
Zero-shot PE-G/14
& 47.4 $\pm$ 5.4 & 60.7 $\pm$ 1.0 & 74.6 $\pm$ 0.9 & 2.2 \\
Global CBM PE-G/14
& 75.8 $\pm$ 7.1 & 77.8 $\pm$ 0.8 & 97.5 $\pm$ 0.4 & 33.6 \\
\midrule
MoTIF ViT-L/14
& 71.0 $\pm$ 6.2 & 76.1 $\pm$ 0.5 & 94.8 $\pm$ 0.5 & 25.8 \\
MoTIF PE-L/14
& 83.2 $\pm$ 6.2 & 81.8 $\pm$ 0.6 & 97.0 $\pm$ 0.3 & 37.3 \\
MoTIF PE-G/14
& \textbf{87.5} $\pm$ 4.9 & \underline{83.0} $\pm$ 0.6 & 98.0 $\pm$ 0.2 & 40.4 \\
MoTIF-ST PE-G/14
& \underline{87.3} $\pm$ 7.1 & 82.1 $\pm$ 1.0 & \underline{98.4} $\pm$ 0.3 & 41.9 \\
\midrule
TSM~\citep{tsm}
& 59.1 & 73.5 & 95.9 & 61.7 \\
NoFrame~\citep{noframe}
& 62.0 & 73.4 & 96.4 & \underline{62.7} \\
VideoMAE V2~\citep{videomae}
& -- & \textbf{88.1} & \textbf{99.6} & \textbf{76.8} \\
\bottomrule
\end{tabular}%
}
\vspace{-1.2em}
\end{wraptable}
\textbf{Baselines.}
Following \citet{dcbm, discover}, we compare MoTIF against zero-shot baselines and supervised reference models.
For fairness, zero-shot predictions are computed at the window level and aggregated by majority voting.
As a supervised interpretable baseline, we train a Global CBM using the same concept vocabulary and backbone features as MoTIF, but collapse the temporal dimension by mean-pooling window-level representations before classification.
This tests whether temporally localized concept modeling adds value beyond a global CBM, while retaining concept-based prediction but removing localized explanations and temporal concept dynamics.
We additionally report strong black-box video models as contextual references rather than strict apples-to-apples baselines.

\subsection{Experimental results}
\textbf{Main comparison.} Table~\ref{tab:perf_comparison} summarizes the main results; the complete backbone sweep and zero-shot baselines are reported in Appendix~\ref{sec:ablation_concept_set}. MoTIF improves over the Global CBM, showing that temporally localized concept activations provide more informative video representations than mean-pooled global concepts.
Accuracy generally increases with backbone capacity, and PE-based backbones outperform CLIP variants at comparable scales. The space--time variant, MoTIF-ST~\citep{timesformer}, is particularly beneficial on SSv2, where classes depend on fine-grained temporal relations. SSv2 remains the most challenging dataset overall due to its abstract relational labels, such as ``putting something onto something.'' On Breakfast, MoTIF improves over the Global CBM by $11$--$18$ percentage points and surpasses two black-box baselines. For HMDB51, UCF101, and SSv2, gains are smaller but consistent, likely because these datasets contain shorter clips where mean pooling is already strong, especially with PE embeddings.

\textbf{Fine-tuned video embedder.} To test whether a fine-tuned embedding model improves performance on SSv2, we trained a LoRA adapter on PE-L/14.
This improves all five MoTIF variants by +2.1 to +10.4 points, with each remaining +2.2 to +10.5 points above its matched adapted Global CBM.
For PE-L/14, the frozen Global CBM and MoTIF achieve 31.3\% and 37.3\%, respectively; after LoRA adaptation, they reach 39.5\% and 44.8\%, preserving a +5.3-point MoTIF advantage.
Although a gap to VideoMAEv2 remains, these results show that task adaptation and the interpretable temporal bottleneck provide complementary gains, substantially narrowing the performance gap while retaining an interpretable concept bottleneck.

\begin{wraptable}{r}{0.55\textwidth}
\vspace{-1.0em}
\centering
\scriptsize
\caption{\textbf{Comparison with DANCE.}
Top-1 accuracy on UCF101, HAA-100, and HAA-500.}
\label{tab:dance_comparison}
\resizebox{\linewidth}{!}{%
\begin{tabular}{@{}llccc@{}}
\toprule
\textbf{Method} & \textbf{Variant / Backbone} & \textbf{UCF101} & \textbf{HAA-100} & \textbf{HAA-500} \\
\midrule
DANCE~\citep{lee2025disentangled} & Baseline w/o interp. & 88.4 & 73.5 & -- \\
DANCE~\citep{lee2025disentangled} & DANCE & 87.5 & 70.7 & -- \\
LF-CBM~\citep{oikarinen2023labelfree} & Disentangled concepts & 85.5 & 66.5 & -- \\
\midrule
MoTIF & ViT-B/32 & 88.5 $\pm$ 0.6 & 61.3 & 55.3 \\
MoTIF & PE-L/14 & 97.0 $\pm$ 0.3 & 87.8 & 80.9 \\
MoTIF & PE-G/14 & \textbf{98.0} $\pm$ 0.2 & \textbf{89.9} & \textbf{84.1} \\
\bottomrule
\end{tabular}%
}
\vspace{-1.2em}
\end{wraptable}

\textbf{MoTIF compared to DANCE.} We further compare MoTIF against DANCE~\citep{lee2025disentangled}, a neighboring baseline for explainable video action recognition. Unlike DANCE, MoTIF explicitly models temporally localized concept activations and their evolution across windows. To clarify this distinction, we report results on UCF101, HAA-100, and HAA-500~\citep{haa500} in Table~\ref{tab:dance_comparison}. For HAA-100, we follow the class merging protocol and use the reported results by \citet{lee2025disentangled}.
The PE-based MoTIF variants outperform DANCE on UCF101 and HAA-100, and we additionally provide results on the larger HAA-500 benchmark. These results suggest that MoTIF scales well while preserving the interpretability benefits of CBMs.

\subsubsection{Ablations}

\label{sec:ablation}
We conduct ablation studies to assess key design choices in MoTIF. Unless stated otherwise, experiments on Breakfast Actions and HMDB51 use CLIP ViT-B/32 and RN/50 backbones, with hyperparameters in Appendix~\ref{app:hyops} and the GPT-5-based concepts listed in Appendix~\ref{ablation_concepts}. This setup spans both transformer-based~\citep{dosovitskiy2020image} and CNN-based~\citep{resnet} visual extractors, as well as datasets with different scale, variability, and action granularity. Additional ablations are provided in Appendix~\ref{app:ablations}.

\begin{figure}[t]
    \centering
    \includegraphics[width=\linewidth]{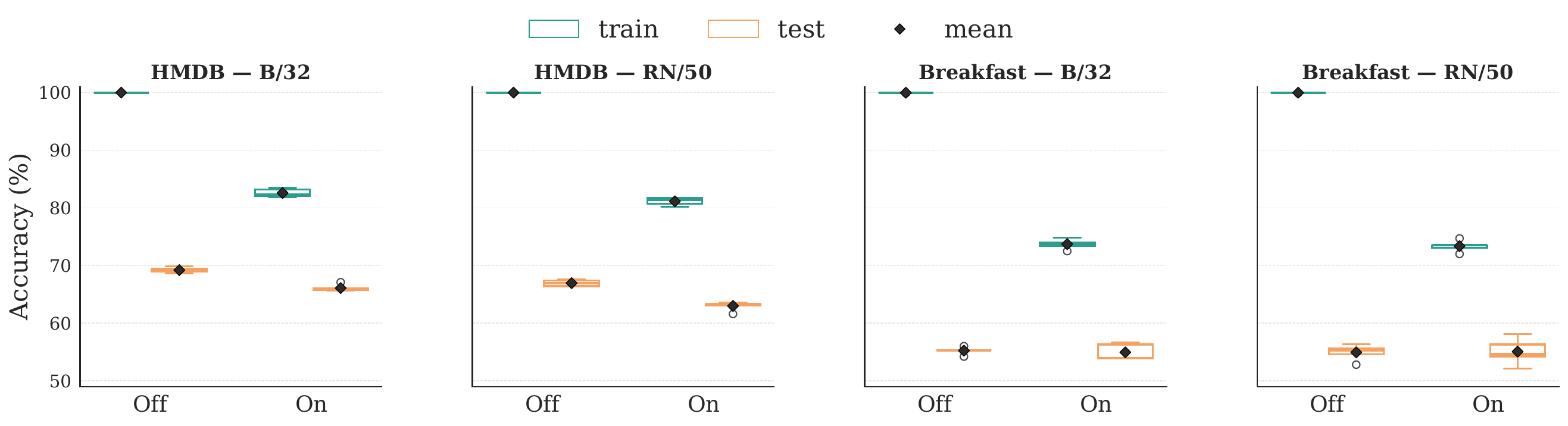}
    \caption{\textbf{Full vs. diagonal attention.}
    Train and test accuracy for diagonal attention (On) and full attention (Off) across five random seeds.}
    \label{fig:diagonal}
\end{figure}

\textbf{Attention variant (full vs.\ diagonal).}
While MoTIF uses per-concept temporal attention, we also evaluate a variant with full multi-head attention.
Figure~\ref{fig:diagonal} compares train and test accuracy for both settings.
Together with the explanation evaluation below, this comparison assesses the impact of cross-concept interactions on both predictive performance and interpretability.
Diagonal attention exposes a clear accuracy--interpretability trade-off: enabling cross-concept attention can improve accuracy on demanding datasets, while diagonal attention supports more direct single-concept attribution, allowing controlled analysis of how temporal interactions affect both prediction and explanation.
On temporally demanding datasets such as SSv2, this trade-off becomes more pronounced, with full attention yielding up to 10.1 percentage points higher test accuracy.
In a blinded LLM-as-judge study of 400 trials restricted to cases where both variants predicted the same class, diagonal-attention explanations were preferred in 67.3\% of comparisons, versus 17.3\% for full attention (15.4\% ties), providing complementary evidence for the explanation side of this trade-off.
We further illustrate the effect on explanations in Section~\ref{full_attention}: when channels are mixed, the most influential dimensions become harder to map back to stable individual concepts, making single-concept attributions less direct.

\textbf{Temporal sensitivity and dynamic concepts.}
A key challenge in video concept learning is preventing models from defaulting to temporally order-invariant reasoning \citep{timesformer}. To assess whether MoTIF captures temporal structure, we evaluate temporal dependence in two complementary ways.
\begin{wraptable}{r}{0.54\textwidth}
\vspace{-1.0em}
\centering
\small
\caption{\textbf{Temporal sensitivity analysis and bottleneck comparison (PE-L/14, GPT-5 concept set, split~1).} Shuffling indicates random permutation of windows at evaluation time; synthetic uses five temporal classes.}
\label{tab:temporal_sensitivity}
\begin{tabular}{lcc}
\toprule
Setting & Basic & Shuffled \\
\midrule
Synthetic (MoTIF) & 87.0 & 21.1 \\
Synthetic (Global CBM) & 35.5 & 35.5 \\
\midrule
Breakfast (MoTIF) & 87.3 & 85.2--86.6 \\
HMDB51 (MoTIF) & 79.9 & 77.1--78.5 \\
UCF101 (MoTIF) & 94.7 & 94.5--94.6 \\
SSv2 (MoTIF) & 30.0 & 26.9--27.4 \\
\bottomrule
\end{tabular}
\vspace{-1.0em}
\end{wraptable}
First, we construct a controlled synthetic benchmark (1{,}989 sequences, matching the number of Breakfast training instances) in which class identity depends exclusively on frame ordering (ascending, descending, U-shaped, inverted-U, and periodic patterns), such that static appearance cues are uninformative by design (for details see Appendix~\ref{app:synthetic_temporal_benchmark}).
MoTIF achieves 87.0\% accuracy on ordered sequences but collapses to 21.1\% under random frame shuffling (chance $\approx 20\%$), yielding a $4.1\times$ performance gap and confirming that, with fixed sinusoidal temporal positional encoding, MoTIF relies on temporal order when required.
Moreover, on this benchmark the static image-level bottleneck baseline (Global CBM) achieves only 35.5\% accuracy, underscoring that temporal modeling is essential when order defines the class.
Second, in Table \ref{tab:temporal_sensitivity}, we apply the same shuffling intervention to real-world datasets.
While some classes admit appearance-based shortcuts, the real-data drops are modest on Breakfast, HMDB51, and UCF101, and largest on SSv2. This suggests that temporal order has the largest measured effect on SSv2, whereas the other benchmarks remain partly solvable from appearance cues alone.
The 2.6--3.1 percentage-point drop on SSv2 is consistent with appearance cues supporting many predictions, as also reflected in the mean-pooled Global CBM results.
The synthetic experiment removes these cues and shows that MoTIF uses temporal order when class identity depends on it.

\subsubsection{Explanations}
\label{sec:explanations}
\begin{figure}[t]
    \centering
    \includegraphics[width=\linewidth]{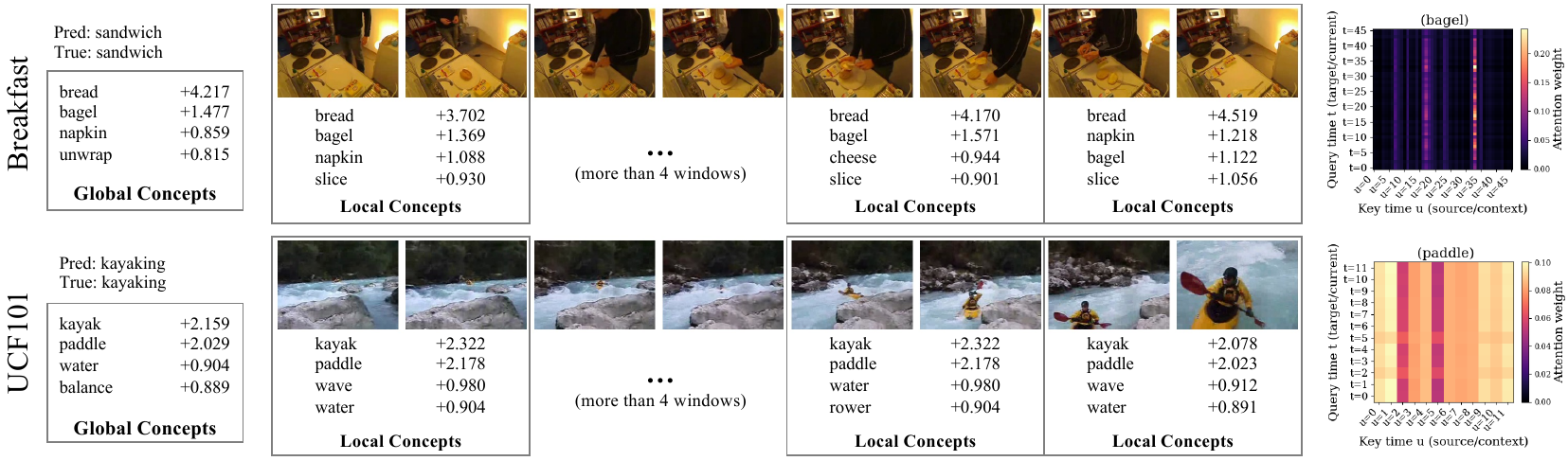}
    \caption{\textbf{MoTIF explanations.} Example videos from Breakfast and UCF101 with correct classifications, illustrating the three explanation modes supported by MoTIF (ViT-L/14).}
    \label{fig:correct_examples}
\end{figure}

A central goal of MoTIF is to provide interpretable insights into the decision-making process of video classification models.
In these examples, the temporal maps are not intended as standalone semantic explanations. Instead, they indicate which temporal windows provide contextual support for updating the queried concept channel, and are interpreted jointly with the local and global concept evidence.
In Figure~\ref{fig:correct_examples}, the first example from Breakfast shows preparing a \textit{sandwich}.
Global and local concept evidence highlight \textit{bread} and \textit{bagel} as important for the prediction.
For the concept \emph{bagel}, the temporal map exhibits strong vertical stripes, indicating that many query frames use the same key frames as contextual support when updating that concept channel.
This suggests that \emph{bagel} is concentrated in a few characteristic windows, while its temporally propagated representation contributes across the sequence.
The second example, from UCF101, depicts \textit{kayaking}.
Here, the most salient concepts are \textit{paddle} and \textit{kayak}, which remain stable over the short clip.
The temporal map for \emph{paddle} is more uniform across time, with slight emphasis at $u=3$ and $u=6$, indicating that contextual support for this concept is distributed across several windows rather than concentrated in a single temporal anchor.
We provide further examples in the supplementary material (Appendix~\ref{app:misses} discusses misclassifications, and Appendix~\ref{full_attention} shows explanations without diagonal attention).

\textbf{Human evaluation of local explanations.} To assess whether local concepts add value beyond global concepts, eight non-author raters evaluated 24 trials, each rated by three raters, yielding 72 ratings.
Each explanation contained eight ranked entries: either eight global concepts or four shared global concepts and four concepts localized to a cited window.
The global-plus-local explanation was preferred in 55.6\% of ratings, compared with 23.6\% for the global-only explanation (20.8\% ties).
With preferences scored as $+1/0/-1$ and bootstrapped by trial, the mean preference was $+0.319$ (95\% CI $[0.083, 0.542]$). This supports the added perceived usefulness of localized concept information.

\textbf{Local--global concept variation.} We next quantify how local explanations differ from the global ranking.
Across Breakfast, HMDB51, SSv2, and UCF101, the mean overlap between each window's top four concepts and the video-level top four is 0.670, 0.859, 0.817, and 0.919, respectively.
Even when the same concepts appear, their rankings can change: same-rank agreement is 0.362, 0.636, 0.597, and 0.727 across these datasets.
Within-video top-four membership variance is highest for Breakfast (0.358), followed by SSv2 (0.185), HMDB51 (0.163), and UCF101 (0.099).
Thus, local explanations vary most in multi-stage Breakfast videos, where different windows can emphasize different parts of an activity.

\subsubsection{Concept interventions}
\label{sec:interventions}

\textbf{Bottleneck-sensitivity analysis.}
We evaluate MoTIF's bottleneck with two intervention types: \textit{destructive interventions}, which test whether predictions depend on sparse concept activations, and \textit{corrective interventions}, which test whether editing these activations can repair errors.
For destructive analysis, Table~\ref{tab:concept_interventions} reports normalized \textit{prediction overlap}, the fraction of post-intervention predictions matching the original MoTIF prediction, with the unperturbed model set to 1.0.
We test global, local-slot, and window interventions.
\textit{Global Top-$k$} zeros the $k$ most influential concept channels across all time steps, while \textit{Global Rand.} removes $k$ random channels as a non-targeted control.
\textit{Local Slot Top-$k$} zeros the $k$ most influential individual concept-window entries, while \textit{Window Top-$k$} zeros all concepts in the $k$ most influential temporal windows and \textit{Window Rand.} zeros random windows.
For corrective interventions, we report \textit{top-1 repair rate} on 30 misclassified instances whose ground-truth class appears in the top-5 logits.
\textit{Global Edit} manually changes up to $k$ selected concepts across all windows, whereas \textit{Local Edit} changes up to $k$ selected concept-window entries.

\begin{table}[h]
\centering
\caption{\textbf{Concept interventions.} Destructive columns report normalized prediction overlap after intervention ($\downarrow$), corrective report top-1 repair rate on misclassified top-5 cases after manual edits ($\uparrow$).}
\vspace{0.5em}
\label{tab:concept_interventions}
\tiny
\resizebox{0.9\linewidth}{!}{%
\begin{tabular}{llccccccc}
\toprule
\textbf{Dataset} & $k$
& \multicolumn{5}{c}{\textbf{Destructive: normalized prediction overlap} $\downarrow$}
& \multicolumn{2}{c}{\textbf{Corrective: top-1 repair rate} $\uparrow$} \\
\cmidrule(lr){3-7}\cmidrule(lr){8-9}
& & \textbf{Global Top-$k$} & \textbf{Global Rand.}
& \textbf{Local Slot Top-$k$} & \textbf{Window Top-$k$} & \textbf{Window Rand.}
& \textbf{Global Edit} & \textbf{Local Edit} \\
\midrule
\multirow{5}{*}{Breakfast}
 & 0 & 1.000 & 1.000 & 1.000 & 1.000 & 1.000 & -- & -- \\
 & 1 & 0.496 & 0.972 & 0.954 & 0.866 & 0.986 & 0.20 & 0.03 \\
 & 2 & 0.229 & 0.950 & 0.933 & 0.775 & 0.977 & 0.47 & 0.10 \\
 & 3 & 0.085 & 0.937 & 0.908 & 0.754 & 0.973 & 0.57 & 0.17 \\
 & 4 & 0.028 & 0.909 & 0.891 & 0.732 & 0.960 & 0.80 & 0.20 \\
\midrule
\multirow{5}{*}{HMDB51}
 & 0 & 1.000 & 1.000 & 1.000 & 1.000 & 1.000 & -- & -- \\
 & 1 & 0.603 & 0.975 & 0.934 & 0.875 & 0.963 & 0.47 & 0.10 \\
 & 2 & 0.374 & 0.959 & 0.892 & 0.801 & 0.947 & 0.60 & 0.23 \\
 & 3 & 0.238 & 0.942 & 0.852 & 0.731 & 0.918 & 0.70 & 0.27 \\
 & 4 & 0.142 & 0.926 & 0.809 & 0.674 & 0.886 & 0.83 & 0.30 \\
\bottomrule
\end{tabular}%
}
\end{table}

Table~\ref{tab:concept_interventions} separates destructive sensitivity from corrective repair.
The unperturbed model has value 1.0. Global top-$k$ removal has the strongest destructive effect: at $k=4$, overlap drops to 0.028 on Breakfast and 0.142 on HMDB51, while random global removal remains above 0.90 on both datasets.
Local slot removal is weaker, which may reflect both temporal redundancy and the smaller perturbation size relative to global concept removal, whereas top-ranked window removal is consistently stronger than random window removal.
For corrective interventions, global edits are substantially more effective than local edits: at $k=4$, global edits repair 80\% of Breakfast cases and 83\% of HMDB51 cases, compared with 20\% and 30\% for local edits.
Repair rates increase monotonically with $k$, suggesting that a small number of globally edited concepts is often sufficient to redirect the prediction to the correct class, while isolated concept-window edits provide more limited corrective control.
Overall, these results show that MoTIF is both sensitive to meaningful concept perturbations and intervenable through targeted concept edits.

\textbf{Qualitative intervention examples.}
The same mechanism also supports instance-level inspection.
In Figure~\ref{fig:start}, zeroing the most influential global concept \textit{bow} changes the prediction from the correct class to \textit{run}; the original correct-class logit is 8.20, while the post-intervention \textit{run} logit is 6.79. Removing windows 1--4, where the bow is handled, instead shifts the prediction to \textit{talk} with logit 6.75.
Conversely, corrective edits can repair errors. For UCF101 video \texttt{v\_ApplyLipstick\_g21\_c01}, MoTIF predicts \textit{ShavingBeard} instead of \textit{ApplyLipstick}; zeroing the misleading concept \textit{barber chair}, which fires because the person is seated, changes the prediction to the correct class.
For SSv2 instance \texttt{100255}, the model predicts \textit{Pretending to scoop something up with something} instead of \textit{Scooping something up with something}; setting the misleading local concept \textit{pouring} to zero only in the first window repairs the prediction.

%% file: sections/related.tex
\section{Related Work}

\textbf{Concept bottleneck models.}
CBMs \citep{cbm_original} predict concepts as intermediate features before the final class prediction \citep{dcbm, labo,havasi2022addressing, chauhan2023interactive, sawada2022concept}. Recent research has focused on automatic concept discovery: DCLIP aligns data with CLIP’s vision--language space \citep{dclip}, LaBo queries large language models for diverse candidate concepts \citep{labo}, and DCBM leverages segmentation foundation models to extract object- and part-level concepts \citep{dcbm}. 
Other works have extended this idea to different modalities. For instance, \citet{proteincbm} adapt the bottleneck principle to protein design, where interpretable biochemical features form the concept space. 
\citet{timeseriescbm} extend CBMs to medical time series data, demonstrating that clinically meaningful temporal features can act as concepts. 
More recently, \citet{llmcbm} apply the CBM framework to language models, where intermediate concepts correspond to interpretable linguistic or semantic units. 
More broadly, MoTIF's separation of information flow across concept channels is related to modular architectures such as Recurrent Independent Mechanisms~\citep{rim}, which also structure computation through partially independent pathways.

CBMs have been widely studied for images, but remain underexplored for sequential data. Prior video extensions either use concepts extracted from video descriptions at a global level~\citep{jeyakumar2022automatic} or disentangle pose-based and textual concepts for comparison~\citep{lee2025disentangled}.
In contrast, MoTIF extends the CBM principle to enable both window-level and global concept attributions for more fine-grained explanations, while also improving performance and remaining flexible with respect to the use of different text–image aligned models and VLMs for concept discovery.


\textbf{Video classification and action recognition.}
Video classification not only assigns a label to an entire sequence but, in the case of action recognition, also requires identifying the actions occurring within it \citep{pareek2021survey}. 
Both tasks have progressed from CNN-based architectures \citep{noframe, tsm, tran2015learning} to transformer-based \citep{cmftransformer} models such as TimeSformer \citep{timesformer} and VideoMAE \citep{tong2022videomae}, which capture long-range dependencies and leverage large-scale self-supervised pre-training. 
MoTIF integrates these approaches by adapting a transformer block for temporal modeling alongside a per-channel scaling operation — distinct from temporal convolution — applied uniformly across all timesteps.

\textbf{Temporal concept modeling.}
While concept-based explanations have been widely studied in the image domain \citep{dseglime, ghorbani2019towards}, only a few works explore temporal dynamics \citep{gulshad2023hierarchical, kowal2024understanding}. 
PCBEAR \citep{pcbear} introduces static and dynamic pose concepts for action recognition, and \citet{tcav} extend TCAV to videos by computing concept importance scores across sequences. 
\citep{Ji_2023_CVPR} explain video models post hoc by automatically discovering spatial-temporal concepts from supervoxels and scoring their importance for 3D ConvNet predictions.
Related object-centric approaches process learned symbolic entities over time for downstream reasoning, including attention over learned object embeddings \citep{objectembeddingsreasoning} and parallelized spatiotemporal slot binding for videos \citep{pstsbvideos}. 
In contrast, MoTIF integrates concept-specific temporal attention directly into the predictive model, enabling reasoning over evolving concepts rather than treating them as fixed inputs or external explanations.

%% file: sections/conclusion.tex
\section{Discussion}
\label{discussion}
\textbf{MoTIF introduces a transformer-based concept bottleneck architecture for temporal data.}
The core contribution of MoTIF is a temporal concept reasoning module that operates on sequences of interpretable concept activations. By combining per-concept temporal self-attention with transformer-style processing, MoTIF models \textit{when} concepts recur and how their temporal patterns contribute to video-level predictions, while preserving explicit concept attribution. Across datasets, this design improves over zero-shot classification, global CBMs that average over windows, and prior video CBM baselines, showing that temporal reasoning can be integrated into concept bottleneck models without discarding concept-level structure.

\textbf{Performance depends on concept quality as well as architecture.}
While MoTIF provides the reasoning structure, its performance depends, as in other CBMs \citep{discover, oikarinen2023labelfree, sawada2022concept}, on the availability of concepts that reflect actions and interactions over time.
VLM-based concept discovery is especially beneficial on temporally demanding datasets such as SSv2, and appendix experiments with SAM3-style visual concepts further suggest that MoTIF is not restricted to textual concept banks, although text remains stronger in our current setup. These findings suggest that architecture and concept discovery are complementary: MoTIF provides the temporal reasoning mechanism, while stronger concept banks improve its ability to handle challenging datasets.

\textbf{MoTIF remains modular while preserving interpretable temporal concept analysis.}
We show that MoTIF can be paired with a range of embedding backbones, including CLIP, SigLIP \citep{siglip}, and video-adapted Perception Encoders, with consistent gains across architectures.
MoTIF-ST further extends the framework to space--time transformers, adding controlled cross-concept interactions for higher expressivity while retaining concept-specific temporal attention. At the same time, this attention has a computational cost: its overhead is small on short clips but becomes substantially larger on long videos (Appendix~\ref{resources}).
Beyond predictive performance, MoTIF's three-level explanation interface---global, local, and temporal---exposes how concepts contribute across time without the entanglement introduced by standard attention mechanisms. This makes it possible to trace errors back to missing or ambiguous temporal concepts rather than opaque feature interactions, helping refine both the concept set and model behavior while maintaining competitive performance.

\textbf{Challenges and future work.} Selecting an appropriate temporal window size remains important, as actions span variable durations and temporal granularities. Adaptive or multi-scale windowing strategies could further improve robustness without sacrificing interpretability. More broadly, performance depends on concept-bank quality and coverage, especially for domain-specific and fine-grained dynamic concepts, suggesting that advances in concept quality may yield larger gains than architectural scaling alone. Although MoTIF works with different concept sources, our experiments do not test whether a vocabulary discovered in one domain transfers to another. Evaluating such transfer is an important direction for future work. Finally, future work may explore selectively enabling concept interactions in a controlled and interpretable manner, reinforcing the value of architectures that make temporal reasoning explicit rather than implicit.

\section{Conclusion}
We introduced MoTIF, a transformer-based concept bottleneck framework for video that combines temporal concept reasoning with VLM-based concept discovery. MoTIF uses diagonal temporal attention to track concepts over time, providing a traceable account of \textit{when} and \textit{which} concepts support predictions. This exposes a controllable accuracy--interpretability trade-off: cross-concept attention can recover additional performance on demanding datasets, while diagonal attention preserves the cleanest concept-level decomposition.
Overall, MoTIF provides a framework for learning, analyzing, intervening on, and explaining temporal concept representations, achieving state-of-the-art performance among CBMs while remaining competitive with black-box video classifiers.

%% file: sections/appendix.tex
\newpage

\section{Implementation details}
\subsection{Hyperparameter settings}
\label{app:hyops}
Unless noted otherwise, all experiments in the main paper follow these default hyperparameters:

\begin{itemize}
    \item \textbf{Training.} 100 epochs, batch size 32, AdamW with learning rate $10^{-3}$ and weight decay $10^{-2}$.
    \item \textbf{Pooling.} Log-sum-exp pooling with fixed temperature $\tau = 1.0$. Although LSE permits tuning of sharpness, we kept $\tau$ constant for unbiased comparisons.
    \item \textbf{Regularization.} Both the $\ell_1$ penalty on classifier weights and the activation sparsity penalty are set to $10^{-3}$.
    \item \textbf{Architecture.} One Transformer layer with per-channel (diagonal) attention; classifier weights constrained to be nonnegative. As stated in Section \ref{sec:method}, the diagonal attention faces the trade-off: interpretability through concept isolation versus representational power.
    Thus, additional depth may yield diminishing returns under this concept-specific design, motivating future work on selectively enabling cross-concept interactions.
    \item \textbf{Data.} Window size of 16 frames per temporal unit. For HMDB51 and Breakfast we report results on split \textit{s1}, and for UCF101 on \textit{testlist01}. Class weighting is applied to mitigate imbalance.
\end{itemize}

Dataset-specific deviations: for SSv2 and UCF101 we use learning rate $10^{-4}$, $\ell_1$ penalty $10^{-4}$, and sparsity penalty $10^{-4}$; for SSv2, the non-negativity constraint on the classifier weights is disabled and the batch size is 128. HMDB51 uses a shorter window size of 8. For Breakfast, the window size is increased to 32 to account for longer videos, and the activation sparsity penalty is set to $10^{-4}$.

All experiments were run with a random seed of 42 for reproducibility. Appendix~\ref{app:ablations} further analyzes the sensitivity to random seed choice.
These values serve as the baseline configuration; modifications for ablation studies are detailed in Section~\ref{app:ablations}.

\subsection{Computation time and complexity}
\label{resources}
Table~\ref{tab:mean-time} reports GPU memory (MB), epoch time (s) and throughput (samples/s) for HMDB51 and Breakfast using the two ablated CLIP backbones. Results are shown for diagonal attention enabled (On) and disabled (Off).

\begin{table}[ht]
\centering
\small
\caption{\textbf{Complexity overview.} GPU memory, epoch time and throughput for RN/50 and B/32 backbones with diagonal attention On/Off.}
\vspace{0.5em}
\label{tab:mean-time}
\begin{tabular}{lllrrr}
\toprule
Dataset & Backbone & Setting & GPU memory (MB) & Epoch time (s) & Samples / s \\
\midrule
\multirow{4}{*}{HMDB51}
  & \multirow{2}{*}{RN/50} & On  & 1723 & 0.91 & 4084 \\
  &                       & Off &  517 & 0.97 & 4063 \\
  & \multirow{2}{*}{B/32}  & On  & 2289 & 0.90 & 4108 \\
  &                       & Off & 1064 & 0.95 & 4071 \\
\midrule
\multirow{4}{*}{Breakfast}
  & \multirow{2}{*}{RN/50} & On  & 10634 & 4.26 & 454 \\
  &                       & Off & 736 & 0.51 & 3461 \\
  & \multirow{2}{*}{B/32}  & On  & 11463 & 4.28 & 445 \\
  &                       & Off & 1565 & 0.53 & 3445 \\
\midrule
\multicolumn{3}{l}{Average $T$ (train set)} & \multicolumn{3}{r}{HMDB51: 12.5 \quad Breakfast: 65.4} \\
\bottomrule
\end{tabular}
\end{table}

For short videos (small $T$) epoch times remain nearly unchanged while GPU memory differs substantially. For longer sequences, as in Breakfast, disabling diagonal attention reduces memory use considerably; the runtime and memory gap grows with sequence length since complexity increases with $T$ (see Section~\ref{sec:method}) if $H \ll C$. The runtime across backbones is similar in magnitude; for example MoTIF with PE-L/14 required 4.3\,s per epoch on Breakfast (Table~\ref{tab:perf_comparison_ablation}). Video embedding time is excluded because embeddings are reusable across models and vary with the embedding backbone. 

\subsection{Backbone integration}
\label{back:integration}

\textbf{Image-based.} Image--text models such as CLIP \citep{clip} and SigLIP \citep{siglip} embed single images rather than videos.
To adapt them, we divide each video into windows of $F$ frames and randomly select one frame per window.
This frame is embedded and serves as the window representation.
Random sampling ensures variation in which part of the window is captured.

\textbf{Video-based.} The Perception Encoder \citep{bolya2025perceptionencoderbestvisual} is explicitly tuned to aggregate information across frames, producing a pooled embedding for each window.
Unlike the image-based backbones, we therefore use multiple frames per window.
In our experiments, we consistently used eight frames per window, except for HMDB51, where the window size itself was eight, so only four frames were sampled.

\subsection{Space-time attention extension}
\label{spaceTime}

While the standard MoTIF architecture applies temporal attention independently for each concept, we extend it with a factorized space-time attention mechanism \citep{timesformer} that enables concept interactions while preserving interpretability.
This design is also related to recent axial-style, modular video binding approaches that preserve slot structure across time \citep{pstsbvideos}.

The \textit{CBMTransformerST} architecture factorizes attention into two sequential components: spatial attention across concepts at each time step, followed by the original per-channel temporal attention.

\textbf{PerTimeSpatialBlock.} This module computes attention across concepts at each time step independently.
Given input features $X \in \mathbb{R}^{B \times T \times C}$, spatial attention produces attention scores $W_s \in \mathbb{R}^{B \times T \times C \times C}$ where $W_{s}[b,t,i,j]$ represents the attention weight from concept $i$ to concept $j$ at time step $t$.

Unlike temporal attention, which uses per-concept projections, spatial attention allows concepts to interact within each temporal frame. To preserve concept interpretability while enabling controlled spatial interaction, we employ three identity-preserving mechanisms:

\begin{itemize}
    \item \textbf{Identity bias}: We add a bias term $\alpha_I = 1.0$ to the diagonal of the attention score matrix, encouraging self-attention and reducing cross-concept mixing.
    \item \textbf{Spatial gating}: The spatial attention output is scaled by a gating factor $\beta_s \in [0,1]$ before being added to the residual connection: $X' = X + \beta_s\cdot \text{SpatialAttn}(X)$. With $\beta_s = 0.1$ (default), the spatial branch contributes a 0.1-scaled residual update, limiting but not eliminating cross-concept mixing.
    \item \textbf{Per-channel FFN}: The feed-forward network in the spatial block uses per-channel convolutions (grouped by $C$), ensuring no cross-concept mixing occurs in the FFN, matching the design of the temporal block.
\end{itemize}

These mechanisms limit cross-concept mixing while retaining a concept-structured representation: the spatial mixing is controlled and identity-preserving, the subsequent per-channel temporal attention maintains diagonal structure, and the final concept activations $Z_{t,c}$ and spatial attention maps $W_s$ enable attribution to individual concepts while revealing their interactions.

\textbf{SpaceTimeBlock.} The block applies spatial attention first to enable concept interactions at each time step, followed by per-channel temporal attention that maintains diagonal structure.

This factorization reduces computational complexity from $\mathcal{O}(T^2 C^2)$ for full space-time attention to $\mathcal{O}(T C^2 + C T^2)$, while producing separate interpretable attention maps for spatial ($W_s \in \mathbb{R}^{B \times T \times C \times C}$) and temporal ($W_t \in \mathbb{R}^{B \times C \times T \times T}$) components.

\textbf{Cross-concept coupling.} To assess how much changing one concept affects the others, we add $\epsilon=0.1$ to one input channel and measure the resulting $\ell_1$ change in the block output.
We define coupling as the average response of the other channels relative to the response of the perturbed channel.
Across 100 SSv2 videos and 16 target channels per video, coupling is 0.14\% for standard MoTIF and 6.82\% for MoTIF-ST, consistent with the spatial block's intended concept interactions.

\textbf{Exemplary explanations.}
MoTIF with CBMTransformerST (MoTIF-ST) applies two sequential attention mechanisms---spatial followed by temporal---yielding two complementary attention matrices for interpretation.
\begin{itemize}
    \item The spatial attention block captures concept--concept interactions within each frame. It indicates which concepts influence a given concept and highlights additional concepts that are relevant at that moment. This provides a structured view of how concepts relate to each other spatially.
    \item The temporal attention block captures when a concept becomes important. It identifies the time steps at which a specific concept contributes most to the final prediction, thereby revealing the temporal structure of the activity.
\end{itemize}

\begin{figure}[t]
\centering
\includegraphics[width=\linewidth]{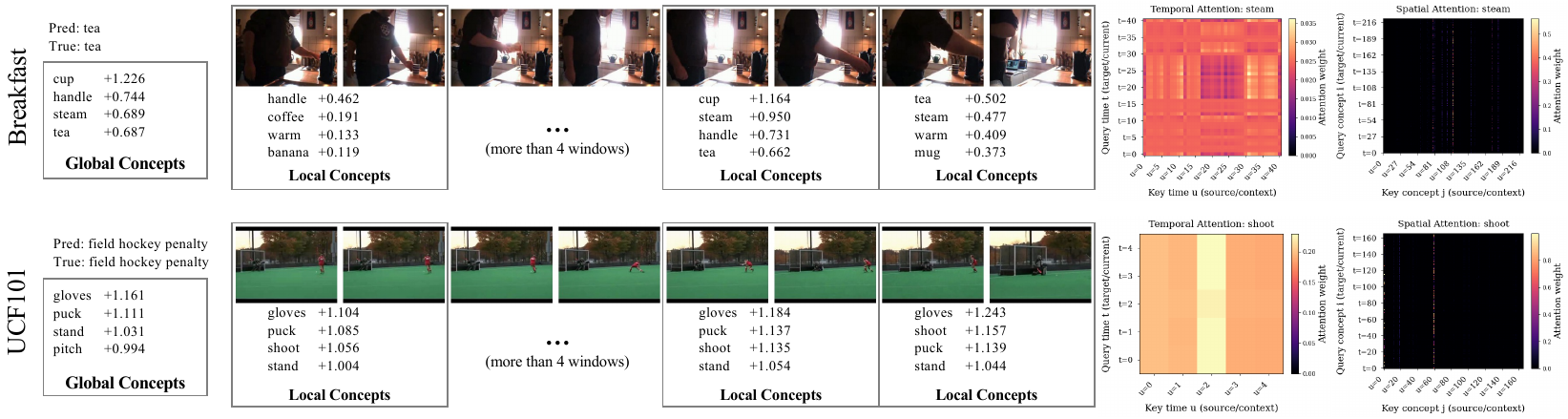}
\caption{\textbf{MoTIF-ST explanations.} Correctly classified examples from Breakfast and UCF101, illustrating MoTIF with a space--time transformer (ViT-L/14) and the corresponding temporal and spatial attention matrices.}
\label{fig:times-former}
\end{figure}

Figure \ref{fig:times-former} shows two examples from Breakfast and UCF101, including their attention matrices and the corresponding explanations.

\section{Additional experiments}
\subsection{Where \textit{motifs} help to understand misclassifications}
\label{app:misses}

\begin{figure}[h]
    \centering
    \includegraphics[width=0.99\linewidth]{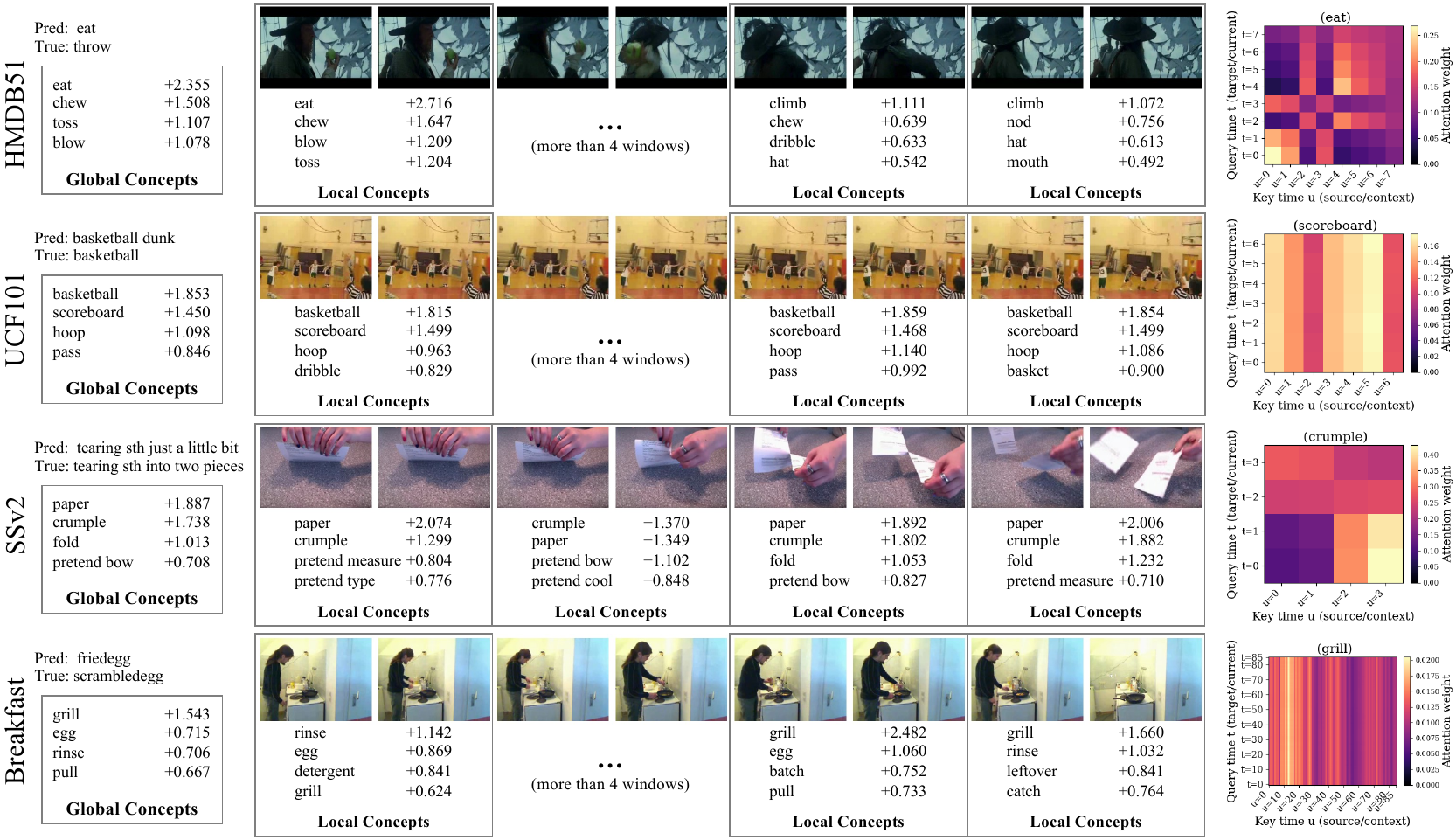}
    \caption{\textbf{MoTIF explanations.} Example videos from all datasets with incorrect classifications.}
    \label{fig:wrong_examples}
\end{figure}

In Figure~\ref{fig:wrong_examples}, we illustrate representative failure cases for each dataset using models trained with ViT-L/14.
All corresponding videos, including the full set of temporal concepts across all windows, are provided in the supplementary material.

The first example, from \emph{Pirates of the Caribbean} (HMDB51), shows Barbossa eating an apple before throwing it away.
Our model predicts \textit{eat}, while the ground truth is \textit{throw}.
Global and local concept attributions reveal that the concept \textit{eat}, triggered by the apple, was most strongly activated.
The corresponding attention map illustrates that early query frames attend to early key frames, indicating that the model anchors the decision on the moment where the apple is clearly visible and eaten.
This concentrated attention explains why the model emphasizes \textit{eat} over \textit{throw}.

In the second example from UCF101, MoTIF detects correct concepts, such as \textit{basketball} and \textit{scoreboard}.
However, these were insufficient to discriminate between the actions \textit{basketball dunk} and \textit{basketball}, leading to misclassification.
Since the background frame remains nearly constant and only the players on the court are moving, the attention map for \textit{scoreboard} shows a uniform distribution across time steps, with no clear temporal anchors.
This indicates that the concept is consistently present.

The third example, from SSv2, highlights the dataset's inherent difficulty.
Unlike the correctly classified case in Figure~\ref{fig:method-overview}, MoTIF predicts \textit{tearing sth just a little bit} instead of the ground truth \textit{tearing sth into two pieces}.
Concept activations focus primarily on hand movements, such as \textit{crumple}, together with the \textit{paper}.
The attention map shows that the concept \textit{crumple} receives the strongest attention from the later key frames ($u=2$, $u=3$) across several query times while other frames receive lower, more diffuse weights.
This diffuse attribution explains why MoTIF captures the general action but fails to resolve the fine-grained distinction required by SSv2.

Finally, in the Breakfast dataset, MoTIF correctly identifies and attends to concepts relevant for \textit{egg}.
Yet, the dataset contains two distinct egg-related actions, \textit{friedegg} and \textit{scrambledegg}, which leads to misclassification.
Nevertheless, all identified concepts correspond to actions of a person cooking with eggs.
Furthermore, the attention map for \textit{grill} reveals a broad and diffuse distribution across many time steps, indicating that the concept is persistently active throughout the sequence rather than concentrated in a few decisive moments.
This persistent but unspecific attention explains why MoTIF captures the general presence of egg-related cooking but fails to distinguish between the two fine-grained classes.

Although MoTIF does not always predict the correct class, its structure makes the reasoning process transparent by decomposing decisions into concepts and their temporal interactions.
In this work, our emphasis was on the architecture rather than on exhaustively optimizing the concept sets.
We expect that future advances in concept extraction for video data will further improve performance, complementing MoTIF's interpretability with stronger predictive accuracy.

\subsection{Diagonal vs.\ full attention}
\label{full_attention}
We illustrate why full attention produces non-interpretable results by revisiting previously shown examples.
In Figure~\ref{fig:start}, replacing diagonal attention with full attention shifts the most important concepts to \emph{frown} (activation 4.796) and \emph{face} (1.282), while all others fall below 0.01 and are omitted.
Although these concepts appear as locally relevant, they do not correspond to the depicted action (class \emph{shoot bow}).
This indicates that concept mixing occurs: full attention entangles channels, creates arbitrary concepts that are not visually apparent, and thus undermines interpretability.

\subsection{Interpreting temporal dependencies}
\label{app:dependencies}
An attention map visualizes how a concept channel distributes its focus across time.
Each entry $W_{c,t,u}$ encodes the attention weight between query time step $t$ (vertical axis) and key/value time step $u$ (horizontal axis).
A bright cell at position $(t,u)$ indicates that the activation of concept $c$ at time $t$ strongly attends to the representation of the same concept at time $u$.
Diagonal patterns suggest that the concept mainly attends to itself at the same or nearby frames, while vertical stripes show that many query frames refer back to the same key frame, indicating the presence of a temporal anchor.
In contrast, diffuse or uniform maps imply that the concept is expressed consistently across time rather than being tied to specific moments.
Thus, by inspecting these maps, one can infer whether a concept is localized, persistent, or temporally linked to particular frames within the sequence.

\subsection{Synthetic temporal data}
\label{app:synthetic_temporal_benchmark}

The synthetic experiment is designed to isolate temporal reasoning by removing all appearance cues. Each ``video'' is a sequence of length $N$ (matching the Breakfast distribution), where each frame is a 1024-dimensional vector. We define five temporal classes: (1) ascending, (2) descending, (3) U-shape, (4) inverted-U, and (5) periodic oscillation. Class identity is determined only by the temporal trajectory of 30\% of the dimensions; the remaining dimensions are filled with noise. Thus, the task can only be solved by learning the temporal order---static appearance information is uninformative by design.

To examine whether the model can learn interpretable temporal structure, we provide 37 artificial temporal pattern concepts. Each concept represents a prototypical analytic temporal signal (ascending, descending, S-curve, exponential growth/decay, chirp, sawtooth, step-up/down, periodic-fast, oscillating-decay, double-peak, plateau, etc.). All embeddings encode temporal shape only, without semantics.

The training setup matches our real pipelines, allowing a direct test of MoTIF's temporal inductive biases. MoTIF achieves 87.0\% accuracy on the original (ordered) sequences but only 21.1\% when frames are randomly permuted (chance 20\%). This $4.1\times$ drop shows that MoTIF relies on temporal ordering rather than static frame statistics. An image-level CBM baseline (``Global CBM''), which collapses all frames into a single vector, reaches only 35.5\%, confirming that methods without temporal modeling cannot solve this task.

%



%

\section{Additional ablations}
\label{app:ablations}

For completeness, we report further ablation studies that complement the main paper (Section~\ref{sec:ablation}).

\begin{figure}[h]
    \centering
    \includegraphics[width=0.9\linewidth]{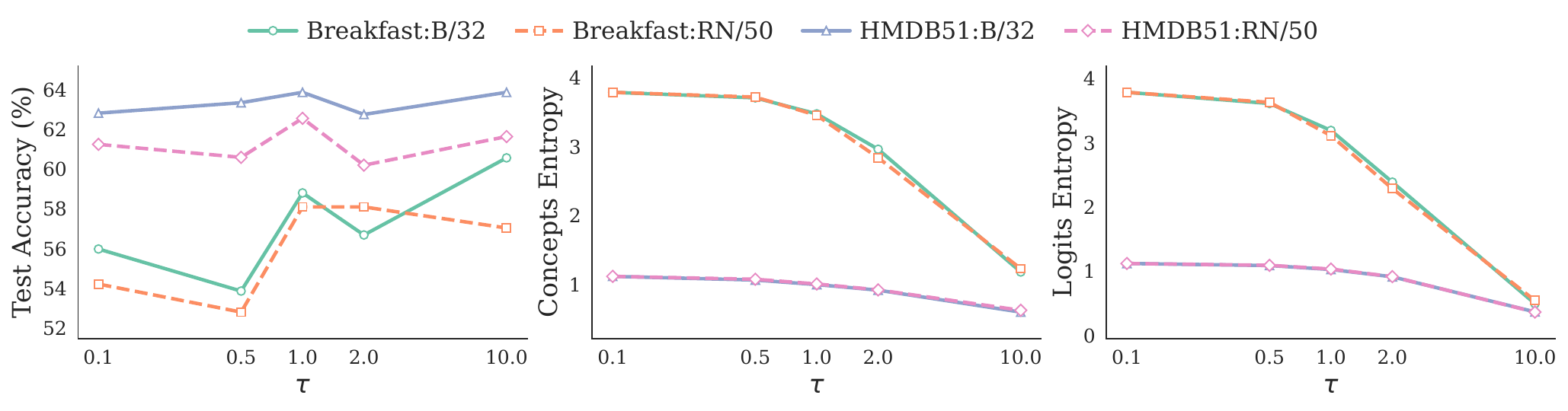}
    \caption{\textbf{Effect of log-sum-exp temperature $\tau$ on accuracy and entropy.}
    Accuracy is stable across $\tau$, while the entropy of the temporal weighting changes systematically with the pooling sharpness.}
    \label{fig:tau}
\end{figure}

\textbf{Temperature $\tau$.}
We vary the log-sum-exp pooling temperature $\tau$ to assess its effect on both accuracy and the sharpness of the time-importance distribution.
Sharpness is quantified by the entropy of the softmax weights, computed either at the concept or at the logits level.
Experiments show (see Figure~\ref{fig:tau}) that accuracy varies only slightly across all tested values of $\tau$, indicating robustness of predictive performance. Under this parameterization, smaller values of $\tau$ yield sharper, lower-entropy temporal attributions, whereas larger values produce smoother, more diffuse weighting across time.
Hence, $\tau$ provides a controllable parameter: small values produce sharp, low-entropy explanations, whereas large values result in smoother, higher-entropy attributions, while accuracy remains relatively stable.
Thus, this parameter can be tuned by the user.

\begin{figure}[t]
    \centering
    \captionsetup[subfigure]{justification=centering}
    \begin{subfigure}[t]{0.45\linewidth}
        \centering
        \includegraphics[width=\linewidth]{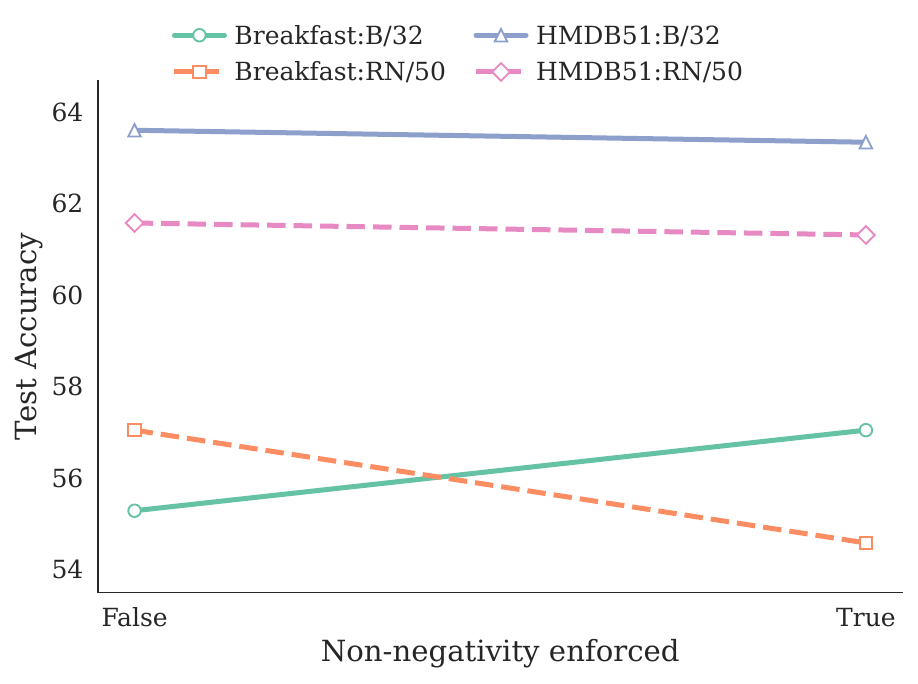}
        \caption{\textbf{Non-negativity.} Test accuracy with and without enforcing non-negativity. }
        \label{fig:nonneg}
    \end{subfigure}\hfill
    \begin{subfigure}[t]{0.45\linewidth}
        \centering
        \includegraphics[width=\linewidth]{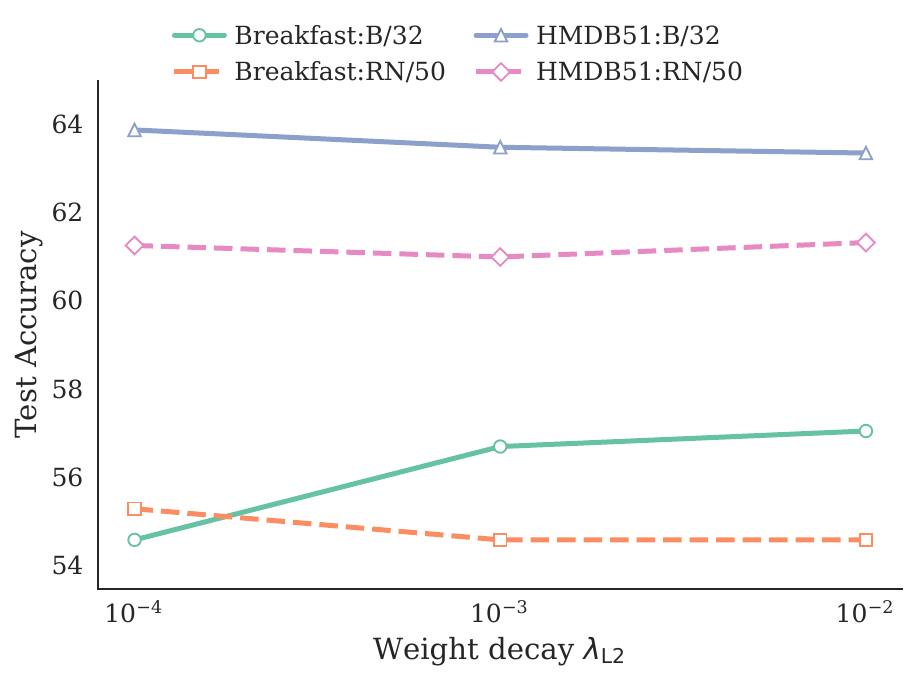}
        \caption{\textbf{Weight decay.} Test accuracy across different weight decays.}
        \label{fig:decay}
    \end{subfigure}
    \caption{\textbf{Architectural choices.} Effects of non-negativity and weight decay.}
    \label{fig:arch-ablations}
\end{figure}

\textbf{Nonnegativity.}
The non-negativity constraint on classifier weights improves interpretability by ensuring class predictions are explained solely by positive concept evidence \citep{zhang2021invertible}.
We enforce non-negativity of classifier weights $W$ by projection after each update.
Results are shown in Figure \ref{fig:nonneg}.
This constraint tends to cause minor accuracy reductions for RN/50 on both datasets, while for B/32 on Breakfast we observe a slight improvement.
Given the small fluctuations inherent in training, these effects should be interpreted as indicative rather than strictly conclusive.
Overall, the effect on test accuracy is negligible.
We therefore enable it for all experiments except SSv2, where modeling negative concepts is required to capture fine-grained actions.

\textbf{Weight decay.}
Figure~\ref{fig:decay} reports test accuracy across different weight decay values for the AdamW optimizer.
Performance remains largely stable, similar to the non-negativity constraint, except for Breakfast using B/32.
We therefore fix the weight decay to $10^{-2}$ for all experiments.

\begin{figure}[h]
    \centering
    \includegraphics[width=0.95\textwidth]{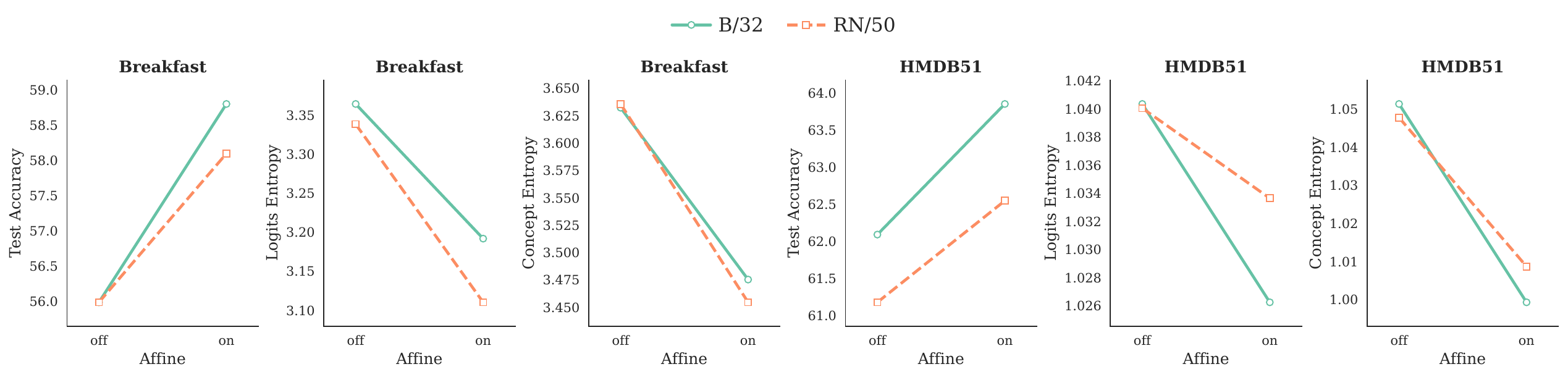}
    \caption{
    \textbf{Effect of the per-concept affine transformation.}
    Accuracy improves marginally, while entropy in both logits and concept activations decreases,
    suggesting that the affine block stabilizes and sharpens the CBM's internal representations.
    }
    \label{fig:affine_effect}
\end{figure}

\textbf{Per-concept affine transformation.} We optionally insert a per-concept affine transformation between the \textit{per-channel temporal self-attention} and the classification head.
While not strictly required, this block sharpens activations by rescaling and shifting each concept dimension before the nonlinearity.
Figure~\ref{fig:affine_effect} shows the effect on Breakfast and HMDB51: enabling the affine transformation yields a slight increase in test accuracy and a consistent reduction in both logits and concept entropy, indicating sharper and more decisive concept activations, especially on the Breakfast dataset.

\begin{figure}[h]
    \centering
    \captionsetup[subfigure]{justification=centering}
    \begin{subfigure}[t]{0.32\linewidth}
        \centering
        \includegraphics[width=0.95\linewidth]{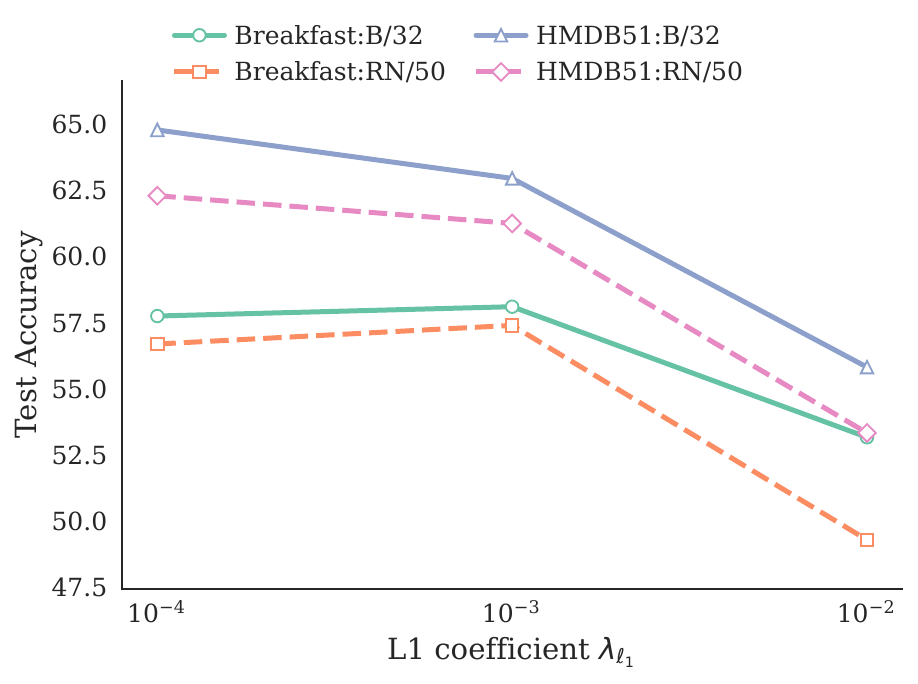}
        \caption{\textbf{Classifier sparsity.} Test accuracy with different $\lambda_{\ell_{1}}$.
        }
        \label{fig:class_sparse}
    \end{subfigure}\hfill
    \begin{subfigure}[t]{0.32\linewidth}
        \centering
        \includegraphics[width=\linewidth]{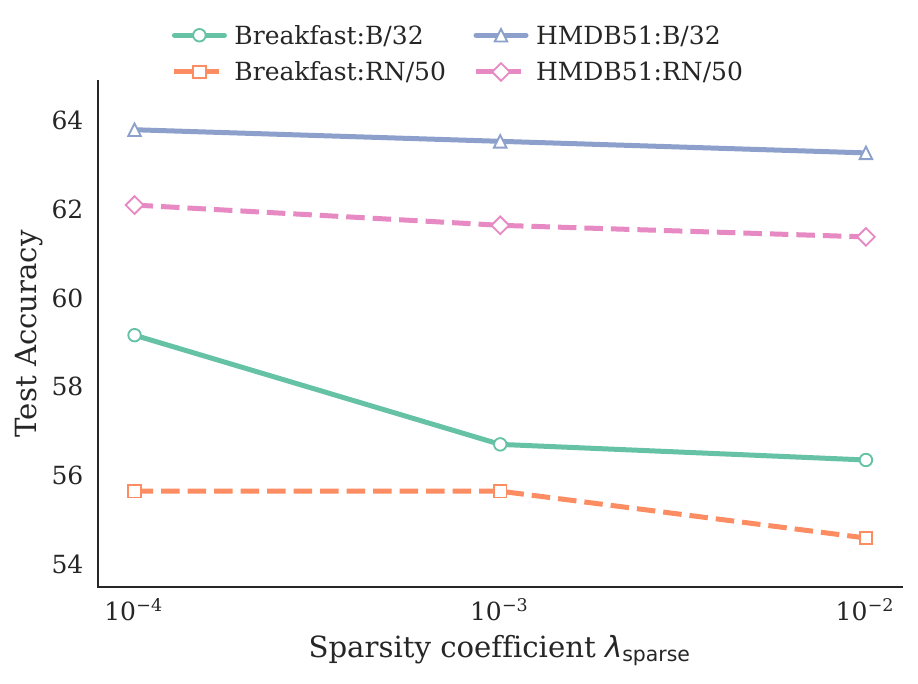}
        \caption{\textbf{Activation sparsity.} Test accuracy across different $\lambda_{\mathrm{sparse}}$.}
        \label{fig:activation_sparse}
    \end{subfigure}\hfill
    \begin{subfigure}[t]{0.32\linewidth}
        \centering
        \includegraphics[width=\linewidth]{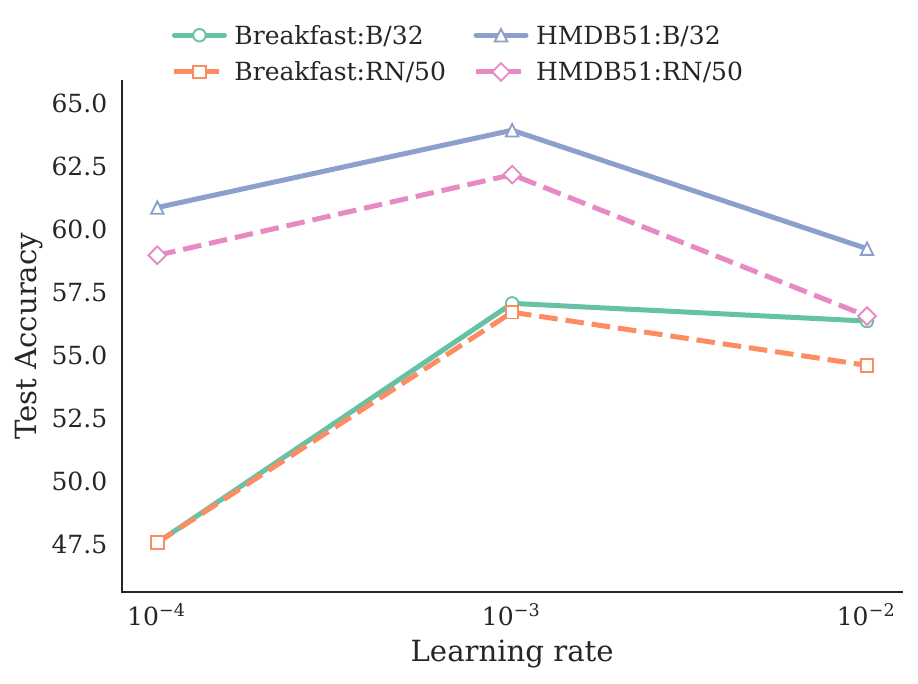}
        \caption{\textbf{Learning rate.} Test accuracy with different learning rates. }
        \label{fig:l1}
    \end{subfigure}
    \caption{\textbf{Architectural choices.} Effects of classifier sparsity, activation sparsity, and learning rate.}
    \label{fig:all_three}
\end{figure}

\textbf{Classifier sparsity.}
The sparsity penalty on classifier weights has a pronounced impact on test accuracy, as shown in Figure~\ref{fig:class_sparse}.
Larger values of $\lambda_{\ell_{1}}$ consistently reduce accuracy.
We therefore set $\lambda_{\ell_{1}} = 10^{-3}$ in most experiments, as it offers a reasonable trade-off between regularization and performance.

\textbf{Activation sparsity.}
The activation sparsity penalty shows little variation in test accuracy (see Figure~\ref{fig:activation_sparse}) across different values of $\lambda_{\mathrm{sparse}}$, except for Breakfast.
We attribute this to the comparatively long video sequences in that dataset.
Accordingly, we use $\lambda_{\mathrm{sparse}} = 10^{-3}$ for all experiments, except for Breakfast, where we set it to $10^{-4}$.

\begin{wrapfigure}{r}{0.4\textwidth}
    \centering
    \vspace{-1.5em}
    \includegraphics[width=\linewidth]{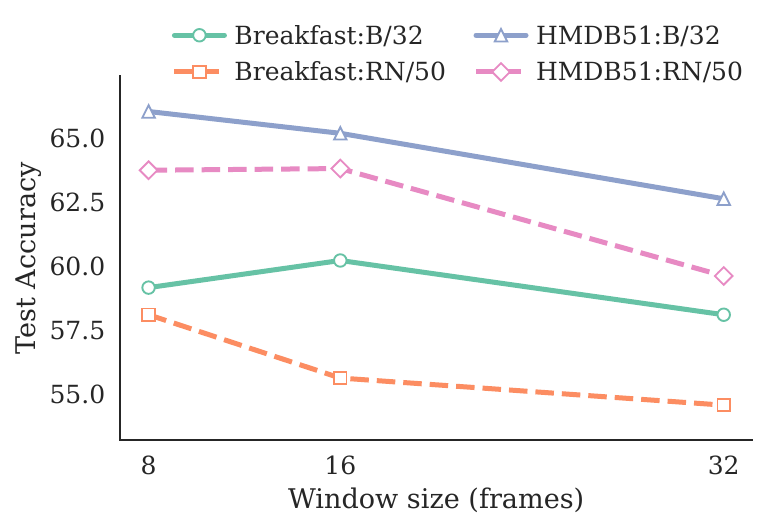}
    \caption{\textbf{Window size influence.}
    Test accuracy across different window sizes.}
    \label{fig:window_size_influence}
    \vspace{-2em}
\end{wrapfigure}
\textbf{Learning rate.}
The learning rate has a strong effect on test accuracy, as shown in Figure~\ref{fig:l1}.
For Breakfast and HMDB51, we set it to $10^{-3}$, while for UCF101 and SSv2, we use $10^{-4}$, which yielded better performance, an effect that was not seen in these ablations.

\textbf{Window size.}
We evaluate MoTIF with varying temporal input lengths to study robustness to sequence duration and efficiency trade-offs.
While increasing the number of frames provides more temporal context, it also raises memory requirements and may not yield consistent accuracy gains.
For comparability, we fix the batch size to $8$ across all ablations, ensuring that changes in performance are solely attributable to window size rather than training dynamics.
The results in Figure~\ref{fig:window_size_influence} highlight that optimal window size depends on both dataset characteristics and backbone choice.

\begin{wrapfigure}{r}{0.6\textwidth}
    \centering
    \vspace{-1.5em}
    \includegraphics[width=0.9\linewidth]{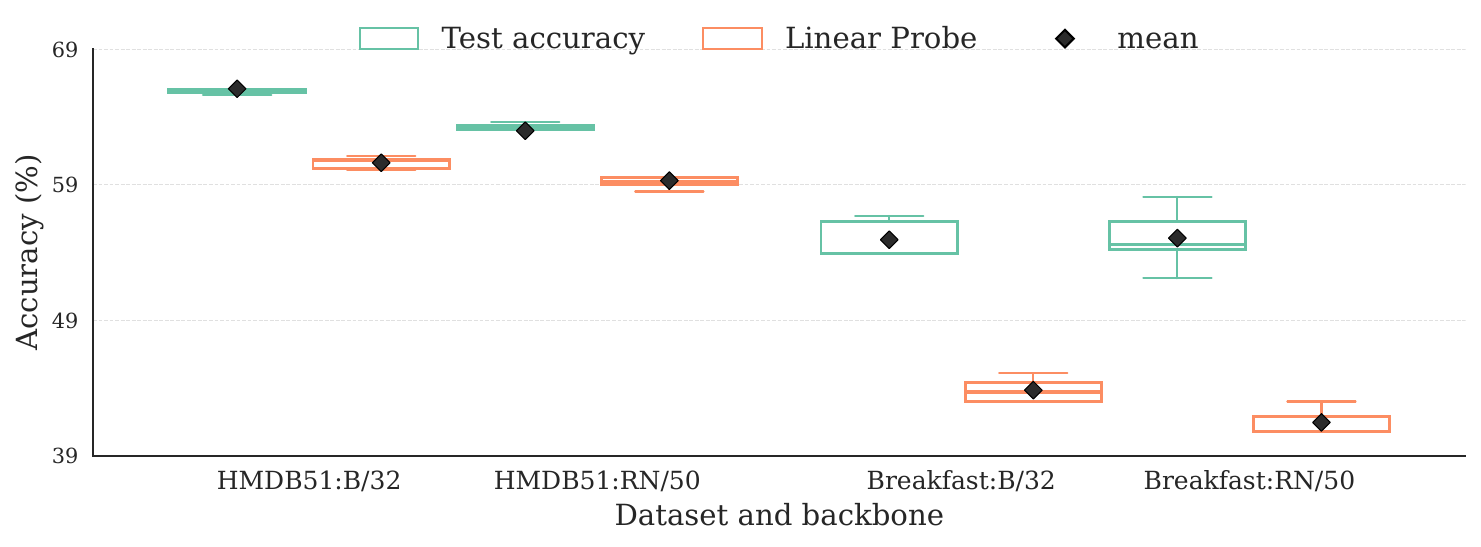}
    \caption{\textbf{Random seed.} Box plot of ablated datasets with five different seeds.}
    \label{fig:random_seed}
\end{wrapfigure}

\textbf{Random seed.}
Since most experiments were run with a fixed seed of 42 for reproducibility, we additionally ablate the effect of varying the random seed (39,40,41,42,43) for both MoTIF and the global CBM.
As Figure~\ref{fig:random_seed} illustrates, the influence of the seed depends on the dataset. For HMDB51 the effect is negligible, whereas for Breakfast --- particularly with the RN/50 backbone --- we observe noticeably higher variance.
Nevertheless, MoTIF consistently outperforms the global CBM.
Moreover, the figure indicates that the reported numbers in Table~\ref{tab:perf_comparison} are conservative: even higher scores are achievable, but we deliberately refrain from reporting maxima and instead provide a representative overview.

Additionally, since embeddings depend on the random choice of a representative image per window (previously fixed at seed 42), we re-embedded each dataset with five seeds.
For each embedding variant we trained and evaluated MoTIF, the global CBM,\ and the zero-shot runs using a fixed training seed of 42 to isolate the effect of image-selection randomness.
Table~\ref{tab:mean-std} reports the mean and standard deviation across embeddings (seeds 39--43). Std values are small overall; the largest observed variability is for Breakfast with RN/50 (std = 1.1).

\begin{table}[ht]
\centering
\small
\caption{\textbf{Random seed on embeddings.} Mean and standard deviation (\%) for methods on HMDB51 and Breakfast with two backbones.}
\vspace{0.5em}
\label{tab:mean-std}
\begin{tabular}{llrr}
\toprule
Dataset (backbone) & Method & Mean (\%) & Std (\%) \\
\midrule
HMDB51 (RN/50) & MoTIF         & 63.8 & 0.3 \\
               & Global CBM  & 60.2 & 0.3 \\
               & Zero Shot     & 29.4 & 0.2 \\
\midrule
HMDB51 (B/32)   & MoTIF         & 66.5 & 0.3 \\
               & Global CBM  & 61.0 & 0.6 \\
               & Zero Shot     & 38.7 & 0.3 \\
\midrule
Breakfast (RN/50) & MoTIF       & 53.2 & 1.1 \\
                 & Global CBM & 41.1 & 1.1 \\
                 & Zero Shot    & 18.1 & 0.7 \\
\midrule
Breakfast (B/32) & MoTIF         & 58.3 & 0.8 \\
               & Global CBM  & 43.9 & 0.6 \\
               & Zero Shot     & 25.8 & 0.2 \\
\bottomrule
\end{tabular}
\end{table}

\subsection{Accuracies on test splits}
\label{testplits}

Tables \ref{tab:breakfast_splits}--\ref{tab:ucf_splits} report Top-1 accuracies (\%) on the test splits. Mean and standard deviation are computed across the shown splits (Breakfast: 4 splits; HMDB51 and UCF101: 3 splits). For each backbone we list the MoTIF result (bold), the corresponding global CBM, and the zero-shot baseline. The reported standard deviation quantifies variability between test splits.

\begin{table}[H]
\small
\centering
\caption{\textbf{Top-1 accuracies (\%) on Breakfast test splits.} Mean and standard deviation across splits.}
\vspace{0.5em}
\label{tab:breakfast_splits}
\begin{tabular}{l|rrrr|rr}
\toprule
Model & s1 & s2 & s3 & s4 & Mean & Std \\
\midrule
\textbf{CLIP RN/50 (MoTIF)}    & 55.6 & 47.4 & 47.0 & 61.2 & 52.8 & 6.9 \\
Global CBM  (RN/50)           & 43.3 & 25.7 & 32.5 & 44.4 & 36.5 & 9.0 \\
Zero-shot (RN/50)              & 17.3 & 19.2 & 16.0 & 21.9 & 18.6 & 2.6 \\
\midrule
\textbf{CLIP B/32 (MoTIF)}         & 56.3 & 44.9 & 51.3 & 61.0 & 53.4 & 6.9 \\
Global CBM  (B/32)                & 42.6 & 27.5 & 31.6 & 47.0 & 37.2 & 9.1 \\
Zero-shot (B/32)                   & 24.6 & 19.8 & 22.0 & 26.4 & 23.2 & 2.9 \\
\midrule
\textbf{CLIP L/14 (MoTIF)}         & 71.5 & 62.3 & 66.7 & 76.8 & 69.3 & 6.2 \\

\textbf{CLIP L/14 (MoTIF-ST)}         & 73.2	&64.8&	65.0	&80.8	&71.0&	7.6 \\
Global CBM  (L/14)                & 61.3 & 44.7 & 48.7 & 66.4 & 55.3 & 10.2 \\
Zero-shot (L/14)                   & 32.4 & 28.3 & 26.5 & 37.0 & 31.1 & 4.7 \\
\midrule
\textbf{SigLIP L/14 (MoTIF)}       & 76.1 & 62.1 & 73.1 & 82.7 & 73.5 & 8.6 \\
Global CBM  (SigLIP L/14)         & 59.9 & 44.7 & 53.3 & 70.6 & 57.1 & 10.9 \\
Zero-shot (SigLIP L/14)            & 28.9 & 18.0 & 20.9 & 26.6 & 23.6 & 5.0 \\
\midrule
\textbf{PE L/14 (MoTIF)}           & 87.3 & 74.7 & 83.1 & 89.4 & 83.6 & 6.5 \\

\textbf{PE L/14 (MoTIF-ST)}           & 86.2	&74.7	&82.3	&91.2	&83.6	&7.0\\
Global CBM  (PE L/14)             & 81.0 & 58.7 & 72.0 & 79.9 & 72.9 & 10.3 \\
Zero-shot (PE L/14)                & 40.5 & 36.6 & 36.8 & 51.6 & 41.4 & 7.0 \\
\bottomrule
\end{tabular}
\end{table}

\begin{table}[H]
\centering
\small
\caption{\textbf{Top-1 accuracies (\%) on HMDB51 test splits.} Mean and standard deviation across splits.}
\vspace{0.5em}
\label{tab:hmdb_splits}
\begin{tabular}{l|rrr|rr}
\toprule
Model & s1 & s2 & s3 & Mean & Std \\
\midrule
\textbf{CLIP RN/50 (MoTIF)}          & 64.1 & 62.3 & 62.1 & 62.8 & 1.1 \\
Global CBM  (RN/50)            & 58.6 & 60.1 & 59.2 & 59.3 & 0.8 \\
Zero-shot (RN/50)               & 29.3 & 30.1 & 30.1 & 29.8 & 0.5 \\
\midrule
\textbf{CLIP B/32 (MoTIF)}          & 65.9 & 66.8 & 63.3 & 65.3 & 1.8 \\
Global CBM  (B/32)                 & 62.0 & 63.0 & 59.8 & 61.6 & 1.6 \\
Zero-shot (B/32)                    & 38.4 & 37.9 & 38.0 & 38.1 & 0.3 \\
\midrule
\textbf{CLIP L/14 (MoTIF)}          & 73.8 & 73.9 & 72.2 & 73.3 & 1.0 \\

\textbf{CLIP L/14 (MoTIF-ST)}          & 75.5	&75.3	&73.6	&	74.8& 1.0 \\
Global CBM  (L/14)                 & 68.5 & 68.8 & 67.9 & 68.4 & 0.5 \\
Zero-shot (L/14)                    & 45.8 & 45.6 & 45.6 & 45.7 & 0.1 \\
\midrule
\textbf{SigLIP L/14 (MoTIF)}        & 74.8 & 74.4 & 70.4 & 73.2 & 2.4 \\
Global CBM  (SigLIP L/14)          & 66.3 & 66.0 & 62.6 & 65.0 & 2.1 \\
Zero-shot (SigLIP L/14)             & 48.4 & 50.0 & 49.5 & 49.3 & 0.8 \\
\midrule
\textbf{PE L/14 (MoTIF)}            & 79.9 & 79.3 & 79.6 & 79.6 & 0.3 \\

\textbf{PE L/14 (MoTIF-ST)}            &79.9&	80.0	&78.8	&	79.6	&0.7 \\
Global CBM  (PE L/14)              & 74.0 & 75.0 & 74.1 & 74.4 & 0.6 \\
Zero-shot (PE L/14)                 & 56.7 & 57.3 & 56.2 & 56.7 & 0.6 \\
\bottomrule
\end{tabular}
\end{table}

\begin{table}[H]
\centering
\small
\caption{\textbf{Top-1 accuracies (\%) on UCF101 test splits.} Mean and standard deviation across splits.}
\vspace{0.5em}
\label{tab:ucf_splits}
\begin{tabular}{l|rrr|rr}
\toprule
Model & s1 & s2 & s3 & Mean & Std \\
\midrule
\textbf{CLIP RN/50 (MoTIF)}    & 82.4 & 82.5 & 83.4 & 82.8 & 0.6 \\
Global CBM  (RN/50)           & 80.7 & 80.0 & 79.3 & 80.0 & 0.7 \\
Zero-shot (RN/50)              & 56.5 & 56.9 & 58.3 & 57.2 & 0.9 \\
\midrule
\textbf{CLIP B/32 (MoTIF)}         & 84.4 & 85.7 & 86.7 & 85.6 & 1.2 \\
Global CBM  (B/32)                & 82.2 & 82.5 & 83.6 & 82.8 & 0.7 \\
Zero-shot (B/32)                   & 59.4 & 60.2 & 60.1 & 59.9 & 0.4 \\
\midrule
\textbf{CLIP L/14 (MoTIF)}         & 92.4 & 93.7 & 93.6 & 93.2 & 0.7 \\

\textbf{CLIP L/14 (MoTIF-ST)}         & 92.9 & 94.4 & 94.0 & 93.8 & 0.8 \\
Global CBM  (L/14)                & 88.8 & 91.0 & 90.3 & 90.0 & 1.1 \\
Zero-shot (L/14)                   & 71.1 & 70.6 & 70.1 & 70.6 & 0.5 \\
\midrule
\textbf{SigLIP L/14 (MoTIF)}       & 93.3 & 93.8 & 94.9 & 94.0 & 0.8 \\
Global CBM  (SigLIP L/14)         & 90.0 & 91.0 & 90.5 & 90.5 & 0.5 \\
Zero-shot (SigLIP L/14)            & 80.0 & 81.9 & 79.2 & 80.4 & 1.4 \\
\midrule
\textbf{PE L/14 (MoTIF)}           & 94.6 & 95.7 & 95.8 & 95.4 & 0.7 \\

\textbf{PE L/14 (MoTIF-ST)}           & 95.7 & 96.4 & 96.9 & 96.3 & 0.6 \\
Global CBM  (PE L/14)             & 94.6 & 93.9 & 95.0 & 94.5 & 0.6 \\
Zero-shot (PE L/14)                & 73.8 & 75.6 & 74.4 & 74.6 & 0.9 \\
\bottomrule
\end{tabular}
\end{table}

\newpage
\subsection{Complete performance comparison}
\label{sec:ablation_concept_set}

Table~\ref{tab:perf_comparison_full} reports the complete evaluation using the VLM-based concept discovery setting.
It compares zero-shot vision-language baselines, Global CBM variants, our MoTIF and MoTIF-ST models, and strong non-interpretable video models across all four datasets.
This table serves as the main performance comparison, while Table~\ref{tab:perf_comparison_ablation} isolates the effect of replacing the VLM-derived concept set with the non-VLM concept set.

\begin{table*}[h]
\centering
\scriptsize
\caption{\textbf{Performance comparison (\% Top-1 accuracy).}
Mean $\pm$ standard deviation on train-test splits on Breakfast Actions, HMDB51, UCF101, and SSv2 with VLM-based concept discovery.}
\label{tab:perf_comparison_full}
\resizebox{0.8\textwidth}{!}{%
\begin{tabular}{lcccc}
\toprule
\textbf{Method} & \textbf{Breakfast} & \textbf{HMDB51} & \textbf{UCF101} & \textbf{SSv2} \\
\midrule
\multicolumn{5}{l}{\textit{Zero-shot}} \\
CLIP-RN/50 \citep{clip}     & 18.6 $\pm$ 2.6 & 29.8 $\pm$ 0.5 & 57.2 $\pm$ 0.9 & 0.8 \\
CLIP-ViT-B/32 \citep{clip}  & 23.2 $\pm$ 2.9 & 38.1 $\pm$ 0.3 & 59.9 $\pm$ 0.4 & 0.9 \\
CLIP-ViT-L/14 \citep{clip}  & 31.1 $\pm$ 4.7 & 45.7 $\pm$ 0.1 & 70.6 $\pm$ 0.5 & 0.9 \\
PE-L/14 \citep{bolya2025perceptionencoderbestvisual}
                           & 41.4 $\pm$ 7.0 & 56.7 $\pm$ 0.6 & 74.6 $\pm$ 0.9 & 2.2 \\
PE-G/14 \citep{bolya2025perceptionencoderbestvisual}
                           & 47.4 $\pm$ 5.4 & 60.7 $\pm$ 1.0 & 74.6 $\pm$ 0.9 & 2.2 \\

\midrule
\multicolumn{5}{l}{\textit{Global CBM}} \\
CLIP-RN/50 \citep{clip}     & 36.9 $\pm$ 7.7 & 61.8 $\pm$ 1.7 & 84.1 $\pm$ 0.8 & 17.0 \\
CLIP-ViT-B/32 \citep{clip}  & 38.3 $\pm$ 9.5 & 62.6 $\pm$ 0.8 & 86.4 $\pm$ 0.6 & 18.2 \\
CLIP-ViT-L/14 \citep{clip}  & 57.0 $\pm$ 8.0 & 71.0 $\pm$ 1.1 & 93.4 $\pm$ 0.7 & 22.0 \\
PE-L/14 \citep{bolya2025perceptionencoderbestvisual}
                           & 72.4 $\pm$ 8.3 & 76.4 $\pm$ 0.8 & 96.3 $\pm$ 0.1 & 31.3 \\
PE-G/14 \citep{bolya2025perceptionencoderbestvisual}
                           & 75.8 $\pm$ 7.1 & 77.8 $\pm$ 0.8 & 97.5 $\pm$ 0.4 & 33.6 \\

\midrule
\multicolumn{5}{l}{\textit{MoTIF (ours)}} \\
MoTIF (RN/50)       & 55.1 $\pm$ 7.1 & 66.2 $\pm$ 0.5 & 86.7 $\pm$ 0.6 & 20.0 \\
MoTIF (ViT-B/32)    & 52.7 $\pm$ 5.8 & 68.5 $\pm$ 1.0 & 88.5 $\pm$ 0.6 & 20.7 \\
MoTIF (ViT-L/14)    & 71.0 $\pm$ 6.2 & 76.1 $\pm$ 0.5 & 94.8 $\pm$ 0.5 & 25.8 \\
MoTIF-ST (ViT-L/14) & 72.6 $\pm$ 6.5 & 75.8 $\pm$ 0.6 & 94.8 $\pm$ 0.4 & 27.7 \\
MoTIF (PE-L/14)     & 83.2 $\pm$ 6.2 & 81.8 $\pm$ 0.6 & 97.0 $\pm$ 0.3 & 37.3 \\
MoTIF-ST (PE-L/14)  & 85.4 $\pm$ 6.3 & 80.8 $\pm$ 1.0 & 97.2 $\pm$ 0.2 & 39.6 \\
MoTIF (PE-G/14)     & \textbf{87.5} $\pm$ 4.9 & \underline{83.0} $\pm$ 0.6 & 98.0 $\pm$ 0.2 & 40.4 \\
MoTIF-ST (PE-G/14)  & \underline{87.3} $\pm$ 7.1 & 82.1 $\pm$ 1.0 & \underline{98.4} $\pm$ 0.3 & 41.9 \\

\midrule
\multicolumn{5}{l}{\textit{Non-interpretable video models}} \\
TSM \citep{tsm}                      & 59.1\footnotemark[1] & 73.5 & 95.9 & 61.7 \\
No frame left behind \citep{noframe} & 62.0\footnotemark[1] & 73.4\footnotemark[1] & \underline{96.4\footnotemark[1]} & \underline{62.7\footnotemark[1]} \\
VideoMAE V2 \citep{videomae}         & -- & \textbf{88.1} & \textbf{99.6} & \textbf{76.8} \\
\bottomrule
\end{tabular}
}
\end{table*}
\footnotetext[1]{from \citep{noframe} with eight clusters and all frames (cumulative)}

Overall, both Global CBM and MoTIF exhibit slightly reduced performance compared to the VLM setting, which we attribute to the smaller and less complete concept set, limiting the amount of task-relevant information captured in the bottleneck.

\begin{table*}[h]
\centering
\caption{\textbf{Performance comparison (\% Top-1 accuracy).}
Mean $\pm$ standard deviation on train-test splits on Breakfast Actions, HMDB51, UCF101, and SSv2 with different CLIP-based backbones with the non-VLM-based concept set. We report seconds per training epoch next to the accuracy scores for all MoTIF variants. }
\label{tab:perf_comparison_ablation}
\resizebox{0.92\textwidth}{!}{%
\begin{tabular}{lcccc}
\toprule
\textbf{Method} & \textbf{Breakfast} & \textbf{HMDB51} & \textbf{UCF101} & \textbf{SSv2} \\
\midrule
\multicolumn{5}{l}{\textit{Zero-shot}} \\
CLIP-RN/50 \citep{clip}       & 18.6 $\pm$ 2.6 & 29.8 $\pm$ 0.5 & 57.2 $\pm$ 0.9 & 0.8 \\
CLIP-ViT-B/32 \citep{clip}   & 23.2 $\pm$ 2.9 & 38.1 $\pm$ 0.3 & 59.9 $\pm$ 0.4 & 0.9 \\
CLIP-ViT-L/14 \citep{clip}   & 31.1 $\pm$ 4.7 & 45.7 $\pm$ 0.1& 70.6 $\pm$ 0.5 & 0.9 \\
SigLIP-L/14 \citep{siglip}   & 23.6 $\pm$ 5.0 & 49.3 $\pm$ 0.8& 80.4 $\pm$ 1.4 & 1.3 \\
PE-L/14 \citep{bolya2025perceptionencoderbestvisual}   & 41.4 $\pm$ 7.0 & 56.7 $\pm$ 0.6& 74.6 $\pm$ 0.9 & 2.2 \\

\midrule
\multicolumn{5}{l}{\textit{Global CBM}} \\
CLIP-RN/50 \citep{clip}       & 36.5 $\pm$ 9.0 & 59.3 $\pm$ 0.8 & 80.0 $\pm$ 0.7& 13.7 \\
CLIP-ViT-B/32 \citep{clip}   & 37.2 $\pm$ 9.1 & 61.6 $\pm$ 1.6& 82.8 $\pm$ 0.7& 15.2 \\
CLIP-ViT-L/14 \citep{clip}   & 55.3 $\pm$ 10.2 & 68.4 $\pm$ 0.5 & 90.0 $\pm$ 1.1& 18.1 \\
SigLIP-L/14 \citep{siglip}   & 57.1 $\pm$ 10.9 & 65.0 $\pm$ 2.1& 90.5 $\pm$ 0.5& 19.6\\
PE-L/14 \citep{bolya2025perceptionencoderbestvisual}   & 72.9 $\pm$ 10.3& 74.4 $\pm$ 0.6& 94.5 $\pm$ 0.6& 25.5 \\

\midrule
\multicolumn{5}{l}{\textit{MoTIF (ours)}} \\
MoTIF (RN/50)       & 52.8 $\pm$ 6.9 (4.2)& 62.8 $\pm$ 1.1 (0.9) & 82.8 $\pm$ 0.6 (1.5)& 16.0 (10.0) \\
MoTIF (ViT-B/32)   & 53.4 $\pm$ 6.9 (4.2)& 65.3 $\pm$ 1.8 (0.9)& 85.6 $\pm$ 1.2 (1.5)& 17.5 (9.9)\\
MoTIF (ViT-L/14) & 69.3 $\pm$ 6.2 (4.3) & 73.3 $\pm$ 1.0 (0.8) & 93.2 $\pm$ 0.7 (1.5) & 20.4 (10.1) \\

MoTIF-ST (ViT-L/14)   & 71.1 $\pm$ 7.7 (7.2) & 74.8 $\pm$ 1.0 (1.8)& 93.8 $\pm$ 0.9 (3.3)& 23.9 (26.8)\\
MoTIF (SigLIP-L/14)  & 73.5 $\pm$ 8.6 (4.2) & 73.2 $\pm$ 2.4 (0.8)& 94.0 $\pm$ 0.8 (1.5) & 22.4 (9.8)\\
MoTIF (PE-L/14)   &\underline{83.6} $\pm$ 6.5 (4.3)& \underline{79.6}$\pm$ 0.3 (0.9)& 95.4 $\pm$ 0.7(1.5)& 30.0 (10.4)\\

MoTIF-ST (PE-L/14)   &\textbf{ 84.1} $\pm$ 6.4 (7.3)& \underline{79.6} $\pm$ 0.7 (1.8)& 96.3 $\pm$ 0.6 (3.3)& 35.1 (26.7)\\

\midrule
\multicolumn{5}{l}{\textit{Non-interpretable video models}} \\
TSM \citep{tsm}                        & 59.1\footnotemark[1] & 73.5 & 95.9 & 61.7 \\
No frame left behind \citep{noframe}   & 62.0\footnotemark[1] & 73.4\footnotemark[1] & \underline{96.4\footnotemark[1]} & \underline{62.7\footnotemark[1]} \\
VideoMAE V2 \citep{videomae}           & --    & \textbf{88.1 } & \textbf{99.6}  & \textbf{76.8 }\\
\bottomrule
\end{tabular}
}
\end{table*}

\newpage
\section{Algorithms}
\label{app:algorithms}

This section summarizes MoTIF's procedures for training, test-time inference, and explanation.
We adopt the notation from the main text: per-window concept activations $Z_{t,c}$, per-time logits $\ell_t$, and log-sum-exp (LSE) pooling with temperature $\tau$ and optional mask $m_t$.

\begin{algorithm}[h]
\caption{Training MoTIF (Moving Temporal Interpretable Framework)}
\label{alg:motif-train}
\begin{algorithmic}[1]
\STATE \textbf{Input:} $\{(X^{(n)}, y^{(n)})\}_{n=1}^N$, concept bank $\mathcal{C}$
\STATE Initialize Transformer parameters; affine $(\gamma,\delta)$; classifier $(W_{\mathrm{cls}},b)$
\FOR{each epoch}
  \FOR{each batch $(X,y)$}
    \STATE \textbf{Temporal modeling:} per-channel temporal self-attention $\to X^{(L)}_{t,c}$
    \STATE \textbf{Affine and Softplus:} $Z_{t,c} \gets \mathrm{Softplus}(\gamma_c X^{(L)}_{t,c} + \delta_c)$
    \STATE \textbf{Classification:} $\ell_t \gets W_{\mathrm{cls}}Z_{t,:}^{\top} + b$
    \STATE \textbf{Pooling:} $\hat{\ell} \gets \mathrm{LSE}_\tau(\{\ell_t\}, m)$
    \STATE \textbf{Loss:} $\mathcal{L} \gets \mathrm{CE}(\hat{\ell}, y) + \lambda_{\ell_1}\|W_{\mathrm{cls}}\|_1
      + \lambda_{\mathrm{sparse}}\frac{1}{(\sum_t m_t)C}\sum_{t,c} m_t |Z_{t,c}|$
    \STATE Update all parameters with AdamW
    \STATE \textbf{Optional:} enforce nonnegativity $W_{\mathrm{cls}} \gets \max(W_{\mathrm{cls}},0)$
  \ENDFOR
\ENDFOR
\STATE \textbf{Output:} trained MoTIF
\end{algorithmic}
\end{algorithm}

\noindent\textbf{Description.}
The training loop refines concept activations with per-channel temporal attention, applies a per-concept affine transformation followed by Softplus, and classifies with a linear head pooled over time via LSE.
The objective combines cross-entropy with sparsity regularizers; $W_{\mathrm{cls}}$ can be projected to enforce nonnegativity.

\begin{algorithm}[h]
\caption{Inference with MoTIF (test-time forward pass)}
\label{alg:motif-infer}
\begin{algorithmic}[1]
\STATE \textbf{Input:} video $x$, trained MoTIF
\STATE Compute concept activations from window embeddings
\STATE Apply per-channel temporal attention; affine + Softplus $\to$ nonnegative $Z_{t,c}$
\STATE Compute per-time logits $\ell_t \gets W_{\mathrm{cls}}Z_{t,:}^{\top} + b$
\STATE Aggregate with LSE pooling: $\hat{\ell} \gets \mathrm{LSE}_\tau(\{\ell_t\}, m)$
\STATE Predict $y^* \gets \arg\max_k \hat{\ell}_k$
\STATE \textbf{Output:} predicted label $y^*$
\end{algorithmic}
\end{algorithm}

\noindent\textbf{Description.}
Inference is a single forward pass: per-window logits are pooled with LSE to produce video-level logits, whose argmax yields the prediction.

\begin{algorithm}[h]
\caption{Explanation with MoTIF (global, local, temporal views)}
\label{alg:motif-explain}
\begin{algorithmic}[1]
\STATE \textbf{Input:} video $x$, class $k$, trained MoTIF
\STATE Run forward pass to obtain $Z_{t,c}$, $\ell_t$, and $\hat{\ell}$
\STATE \textbf{Per-time contributions:} $\mathbf{c}_t^{(k)} \gets Z_{t,:}\odot W_{\mathrm{cls},k,:}$
\STATE \hspace{1.65em} $s_t^{(k)} \gets \sum_c c_{t,c}^{(k)} + b_k$
\STATE \textbf{Temporal importance:} $\pi_t^{(k)} \gets \mathrm{softmax}\!\big(s_t^{(k)}/\tau\big)$ over valid $t$ (mask $m_t$)
\STATE \textbf{Global attribution:} $\bar{\mathbf{c}}^{(k)} \gets \sum_t \pi_t^{(k)}\,\mathbf{c}_t^{(k)}$
\STATE \textbf{Output:} $\big(\bar{\mathbf{c}}^{(k)}, \{\pi_t^{(k)}\}_t, \text{attention maps}\big)$
\end{algorithmic}
\end{algorithm}

\noindent\textbf{Description.}
Explanations derive per-concept attribution scores and reweight them by a time-importance distribution that mirrors LSE pooling.
This yields three complementary views: (1) global concepts via $\bar{\mathbf{c}}^{(k)}$, (2) local concepts at high-weight windows (large $\pi_t^{(k)}$), and (3) temporal dependencies from per-concept attention maps.

\begin{algorithm}[h]
\caption{VLM-based concept discovery for video (objects + actions)}
\label{alg:agentic-concepts}
\begin{algorithmic}[1]
\STATE \textbf{Input:} training set $\mathcal{D}_{\text{train}}$, VLM $A_{\mathrm{vlm}}$, windows $T$, frames/window $l$, max videos/class $n$, similarity threshold $\rho$
\STATE \textbf{Output:} filtered concept set $\mathcal{C}$

\STATE Initialize candidate pool $\mathcal{C}_{\text{raw}} \gets [\,]$
\FOR{each class $y$}
    \STATE Sample up to $n$ videos $\{x\}$ of class $y$ from $\mathcal{D}_{\text{train}}$
    \FOR{each video $x$}
        \STATE Split $x$ into $T$ windows and sample $l$ frames per window
        \FOR{each window $w_t$}
            \STATE $(\mathcal{C}^{\mathrm{obj}}_{w_t},\mathcal{C}^{\mathrm{act}}_{w_t}) \gets A_{\mathrm{vlm}}(w_t,y)$
            \STATE Append both concept sets to $\mathcal{C}_{\text{raw}}$
        \ENDFOR
    \ENDFOR
\ENDFOR

\STATE Embed all concepts using text encoder $\psi(\cdot)$
\STATE Filter $\mathcal{C}_{\text{raw}}$ by cosine similarity threshold $\rho$ to obtain $\mathcal{C}$

\STATE \textbf{return} $\mathcal{C}$
\end{algorithmic}
\end{algorithm}

\noindent\textbf{Description.}
VLM-based concept discovery decomposes videos into short temporal windows and uses a vision--language model to extract textual object and action concepts from sampled frames.
Concepts are aggregated across windows and videos into a shared candidate pool and filtered by similarity to remove redundant entries.
This yields a concept set constructed from training data with weak supervision from class labels.



\newpage
\section{Concept sets}
\label{app:concepts}
\subsection{Concept discovery with VLM approach}

This section evaluates concept-bank quality from several complementary but mostly indirect perspectives. Rather than directly annotating the semantic correctness of every discovered concept or per-window concept assignment, we study how downstream behavior changes when varying the discovery model, concept filtering threshold, concept source, and concept vocabulary. We view these analyses as targeted evidence about concept coverage, diversity, and temporal relevance.

\textbf{Ablation on VLM backbones.}
To study the impact of the underlying vision--language model, we repeat the concept discovery procedure using different Qwen~3 backbones (4B, 8B, and 30B parameters) on Breakfast and HMDB51.
Table~\ref{tab:qwen_backbones} reports the resulting mean classification accuracy together with the average number of discovered concepts.
Note that the number of concepts can vary across splits, as concepts are extracted solely from the training videos.

\begin{table}[h]
\centering
\caption{\textbf{Ablation on Qwen~3 backbones for VLM-based concept discovery.}
Mean Top-1 accuracy and average number of discovered concepts (train split only),
evaluated with different CLIP vision backbones.}
\vspace{0.5em}
\label{tab:qwen_backbones}
\small
\begin{tabular}{llcccc}
\toprule
\textbf{Qwen} & \textbf{Vision} &
\multicolumn{2}{c}{\textbf{Breakfast}} &
\multicolumn{2}{c}{\textbf{HMDB51}} \\
\cmidrule(lr){3-4} \cmidrule(lr){5-6}
& \textbf{Backbone} & Acc. (\%) & \#Concepts & Acc. (\%) & \#Concepts \\
\midrule
Qwen~3~4B  & RN/50     & 55.2 & 338 & 66.2 & 793 \\
           & ViT-B/32  & 53.3 & 240 & 68.6 & 618 \\
\midrule
Qwen~3~8B  & RN/50     & 54.5 & 330 & 66.8 & 872 \\
           & ViT-B/32  & 54.1 & 238 & 68.6 & 662 \\
\midrule
Qwen~3~30B & RN/50     & 55.1 & 258 & 66.2 & 657 \\
           & ViT-B/32  & 52.7 & 193 & 68.5 & 511 \\
\bottomrule
\end{tabular}
\end{table}

\textbf{Ablation on similarity threshold.}
We further ablate the effect of the similarity threshold used for concept filtering on Qwen~3 30B.
While our main experiments use a threshold of 0.9, Table~\ref{tab:similarity_threshold} reports results for thresholds of 0.8 and 0.7, together with the corresponding concept counts.

\begin{table}[h]
\centering
\caption{\textbf{Ablation on concept similarity threshold.}
Impact on concept count and downstream accuracy for different vision backbones.}
\vspace{0.5em}
\label{tab:similarity_threshold}
\small
\begin{tabular}{llcccc}
\toprule
\textbf{Threshold} & \textbf{Vision Backbone} &
\multicolumn{2}{c}{\textbf{Breakfast}} &
\multicolumn{2}{c}{\textbf{HMDB51}} \\
\cmidrule(lr){3-4} \cmidrule(lr){5-6}
& & Acc. (\%) & \#Concepts & Acc. (\%) & \#Concepts \\
\midrule
0.7 & RN/50    & 39.5 & 24 & 56.1 & 51 \\
    & ViT-B/32 & 29.9 & 11 & 49.4 & 27 \\
\midrule
0.8 & RN/50    & 51.5 & 88 & 63.8 & 217 \\
    & ViT-B/32 & 49.1 & 52 & 64.3 & 128 \\
\midrule
0.9 & RN/50    & 55.1 & 258 & 66.2 & 657 \\
    & ViT-B/32 & 52.7 & 193 & 68.5 & 511 \\
\bottomrule
\end{tabular}
\end{table}

For a threshold of 0.9, the number of filtered concepts ranges from 1,748--2,325 on \textsc{SSv2}, 481--1,217 on \textsc{HMDB51}, 186--366 on \textsc{Breakfast}, and 729--2,075 on \textsc{UCF101}.
The large variance is primarily driven by the embedding model and the size of the training splits.
In this ablation, larger concept sets are associated with improved performance, suggesting that increased concept diversity can yield a more complete bottleneck representation.

\textbf{Visual concepts via SAM3.}
Motivated by DCBM's use of segmentation-derived visual concepts \citep{dcbm}, we additionally replace the textual concept bank with visual concepts obtained from a promptable segmentation model (SAM3) \citep{carion2026sam}. In this setting, the discovery pipeline proposes visual regions and short visual snippets rather than text labels, which are then embedded and used as the bottleneck vocabulary. Table~\ref{tab:sam3_visual_concepts} shows that this visual-concept variant remains viable across datasets and backbones, although it underperforms the textual concept bank in our current setup. We therefore treat it as a concept-quality robustness check showing that MoTIF is compatible with DCBM-style visual concept sources rather than as a new main result.

\begin{table}[t]
\centering
\caption{\textbf{Visual-concept robustness check with SAM3-style concepts.} Mean $\pm$ standard deviation across official splits for Breakfast, HMDB51, and UCF101; SSv2 is reported on its single official validation split.}
\vspace{0.5em}
\label{tab:sam3_visual_concepts}
\small
\begin{tabular}{lcccc}
\toprule
\textbf{Backbone} & \textbf{Breakfast} & \textbf{HMDB51} & \textbf{UCF101} & \textbf{SSv2} \\
\midrule
RN/50    & 49.4 $\pm$ 3.8 & 57.5 $\pm$ 1.1 & 81.3 $\pm$ 0.6 & 16.5 \\
ViT-B/32 & 50.6 $\pm$ 6.8 & 60.5 $\pm$ 1.5 & 84.6 $\pm$ 0.8 & 18.1 \\
ViT-L/14 & 66.6 $\pm$ 5.3 & 67.2 $\pm$ 1.1 & 92.1 $\pm$ 1.1 & 22.2 \\
PE-L/14  & 80.7 $\pm$ 5.3 & 74.5 $\pm$ 1.5 & 96.7 $\pm$ 0.2 & 32.5 \\
\bottomrule
\end{tabular}
\end{table}

\textbf{Ablation of concept types.}
To more directly analyze the role of different concept types, we conducted an additional ablation in Table \ref{tab:concept_type_ablation} that separates the concept sets used by MoTIF into \textit{object concepts}, \textit{action concepts}, and their \textit{combination}, and evaluates them across the main datasets.
While this remains a downstream evaluation rather than a direct semantic audit, it provides a more targeted analysis than the previous results, as it isolates the contribution of each concept type and, in particular, tests whether action concepts yield measurable benefits beyond object concepts.

\begin{table}[h]
\centering
\caption{\textbf{Concept types ablation.} Performance of MoTIF when using different concept types across datasets and backbones.}
\vspace{0.5em}
\small
\begin{tabular}{llcccc}
\toprule
Backbone & Concepts & Breakfast & HMDB51 & UCF101 & SSv2 \\
\midrule
L/14    & Object            & $70.9\pm7.2$ & $73.4\pm1.4$ & $94.5\pm0.7$ & $24.1$ \\
L/14    & Action            & $71.0\pm7.6$ & $77.1\pm0.9$ & $95.0\pm0.4$ & $24.6$ \\
L/14    & Object + Action   & $71.0\pm6.2$ & $76.1\pm0.5$ & $94.8\pm0.5$ & $25.8$ \\
\midrule
PE-L/14 & Object            & $83.6\pm7.4$ & $80.0\pm1.5$ & $96.8\pm0.3$ & $33.8$ \\
PE-L/14 & Action            & $85.1\pm5.8$ & $81.3\pm0.5$ & $97.0\pm0.5$ & $36.4$ \\
PE-L/14 & Object + Action   & $83.2\pm6.2$ & $81.8\pm0.6$ & $97.0\pm0.3$ & $37.3$ \\
\bottomrule
\end{tabular}
\label{tab:concept_type_ablation}
\end{table}

The results reveal a consistent pattern: action concepts are particularly beneficial on the more temporally demanding benchmarks.
This is most evident on HMDB51 and SSv2, where action concepts outperform object concepts for both backbones.
For example, with PE-L/14 on SSv2, performance increases from $33.8$ with object concepts to $36.4$ with action concepts, and further to $37.3$ when object and action concepts are combined.
A similar trend can be observed on HMDB51, where the improvement from object to action concepts is also consistent across both backbones.

These findings are in line with the intended role of action concepts: they appear to capture information that goes beyond static object cues and is particularly useful when temporal structure matters.
At the same time, combining object and action concepts gives the best results on SSv2 and matches or approaches the best single type elsewhere, suggesting that both concept types provide complementary information rather than redundant signals.
This complementarity is further encouraged by our concept filtering procedure, which removes overly similar concepts and thus promotes a more diverse concept set.

\newpage

\subsection{LLM concepts}
\label{ablation_concepts}
In Table \ref{tab:concepts}, we show the number and kind of concepts used for the construction of MoTIF for each dataset. The number and kind of concepts vary for each dataset, since we asked the LLM to create domain-specific concepts that are useful for the downstream classification task. As the table indicates, the number of concepts is lower than for the VLM-based concept sets.

\begin{table}[ht]
\centering
\caption{\textbf{MoTIF GPT-5 concepts.} The textual concepts utilized for the ablation experiments (Section~\ref{sec:ablation}); the main results use the VLM-discovered concept sets.}
\vspace{0.5em}
\tiny
\begin{tabular}{lc p{11cm}}
\toprule
\textbf{Set} & \textbf{Count} & \textbf{Concepts} \\
\midrule
Breakfast & 223 & add, adjust, apple, arrange, assemble, avocado, bacon, bagel, bake, balance, banana, batter, beat, bin, blend, blender, blow, boil, bottle, bowl, bread, brew, brush, burner, butter, button, carry, carton, catch, cereal, chair, cheese, chop, cinnamon, clap, close, coffee, colander, comb, container, cook, cookbook, cool, core, counter, cover, crack, croissant, cucumber, cup, cupboard, cut, cuttingboard, detergent, dish, drag, drain, drizzle, drop, dry, egg, faucet, fill, flame, flip, fold, fork, freezer, fridge, froth, frown, fruit, fry, garbage, gesture, granola, grate, grater, grill, grind, ham, handle, heat, herb, hide, honey, hood, ice, ingredient, insert, jar, juice, kettle, knife, knob, knock, ladle, laugh, leftover, lid, mash, measure, measuringcup, measuringspoon, milk, mix, mug, napkin, nod, onion, open, orange, oven, ovenmitt, pack, package, pan, pantry, pastry, peel, peeler, pick, pinch, pit, place, plate, plug, point, poke, pour, preheat, press, pull, push, put, reach, recipe, recycle, release, remove, reveal, rinse, roll, rotate, sausage, scale, scoop, scramble, scrub, seal, serve, serving, set, shake, shave, sieve, sink, sip, sit, slice, slide, smile, snap, soap, socket, sort, spatula, spin, sponge, spoon, spread, sprinkle, squeeze, stack, stand, start, steam, steep, stir, stirrer, stool, stop, stove, strawberry, sugar, switch, syrup, table, take, tamp, tap, taste, tea, thermometer, throw, tie, tilt, timer, toast, tomato, tongs, toss, towel, tray, turn, twist, uncover, unfold, unscrew, unstack, untie, unwrap, warm, wash, waste, water, wave, whisk, wipe, wring, yogurt, zest, zip \\
\midrule
UCF101& 166 & aim, archer, archery, arena, arrow, athlete, balance, ball, bar, barbell, baseball, basket, basketball, bat, beam, bicycle, block, bounce, bow, bowl, boxing, breakdance, canoe, cap, catch, clap, climb, club, coach, control, court, cricket, curl, dance, dancer, deadlift, dismount, dive, dodge, dribble, dumbbell, enter, field, fight, flip, floor, frisbee, gallop, gloves, goal, goalpost, grab, grapple, grind, gun, gym, gymnast, handstand, hang, helmet, hit, hockey, hook, hoop, horse, hurdle, ice, instrument, jab, jersey, jump, kayak, kick, ladder, lane, lift, mat, microphone, mount, music, net, netting, opponent, pad, paddle, parry, pass, pedal, perform, pitch, platform, player, pool, press, puck, pull, push, racket, rail, raise, referee, reins, release, reload, ride, ring, rope, row, rower, rugby, run, sand, scoreboard, serve, sheet, shoot, shooter, sit, skateboard, skateboarder, skater, ski, skip, skis, smash, snow, snowboard, snowboarder, soccer, spike, spin, splash, sprint, squat, stadium, stage, stand, start, steer, stick, stop, strike, surf, surfboard, surfer, swim, swimmer, swing, sword, target, teammate, throw, timer, track, trampoline, tuck, turn, uniform, uppercut, volleyball, walk, wall, water, wave, wrestle, yoga \\
\midrule
HMDB51 & 150 & apply, around, backward, balance, ball, baseball, basketball, bat, bend, bicycle, block, blow, bottle, bounce, bow, brake, brush, button, carry, cartwheel, catch, chair, chew, climb, close, comb, crawl, cross, crouch, cup, dismount, dive, door, down, drag, dribble, drink, drop, eat, enter, exit, face, fall, fight, finish, flip, float, frisbee, from, frown, gallop, grab, hair, hand, hands, handstand, hat, head, headstand, high, hit, hop, horse, hug, jacket, jog, juggle, jump, kick, kiss, knock, laugh, leap, left, leg, lie, lift, line, look, low, makeup, mount, mouth, nod, object, off, on, open, pedal, point, pull, punch, push, put, racket, reach, release, ride, right, roll, room, run, serve, shake, shave, shirt, shoelace, shoot, sing, sip, sit, skate, skateboard, ski, sled, sleep, slide, smile, snowboard, somersault, spin, sprint, stand, start, steer, stretch, surface, swim, swing, sword, take, talk, teeth, tennis, throw, tie, toss, touch, turn, untie, up, utensils, wake, walk, wash, wave, with, words, yawn, zip
 \\
\midrule
SSv2 & 284 & accelerate, apple, arm, assemble, background, backpack, bag, balance, ball, banana, bend, bite, blow, book, bottle, bottom, bounce, bow, bowl, box, break, broken, can, cap, carrot, carry, catch, chair, chew, chop, chopstick, clap, clean, click, climb, close, closeable, cold, connect, container, cough, cover, crawl, crouch, crumple, cry, cucumber, cup, cut, dance, decelerate, dirty, disassemble, disconnect, door, downward, drag, drag mouse, draw, drink, drinkable, drop, dry, durable, eat, edible, empty, erase, face, fall, fasten, fill, finger, fixed, flatten, flip, floor, fold, fork, fragile, frown, fruit, full, gather, get up, grape, hand, heavy, hide, hold, hop, hot, insert, inside, juggle, jump, key, keyboard, kneel, knife, knock, laptop, laugh, lean, left, lid, lift, light, lock, loosen, mix, mouse, nod, object, open, openable, orange, other, outside, paint, paper, peel, pen, pencil, person, phone, plate, plug, point, pour, pourable, press, pretend to balance, pretend to block, pretend to bow, pretend to catch, pretend to catch fish, pretend to clap, pretend to clean, pretend to climb, pretend to close, pretend to cook, pretend to dance, pretend to dodge, pretend to draw, pretend to dribble, pretend to drink, pretend to drive, pretend to eat, pretend to fall, pretend to fire gun, pretend to honk, pretend to hug, pretend to jump rope, pretend to kick, pretend to kiss, pretend to load gun, pretend to lock, pretend to look around, pretend to measure, pretend to open, pretend to paddle, pretend to paint, pretend to play drums, pretend to play guitar, pretend to play piano, pretend to point, pretend to pour, pretend to pull, pretend to punch, pretend to push, pretend to read, pretend to row, pretend to salute, pretend to scroll, pretend to search, pretend to serve, pretend to shake hands, pretend to shoot arrow, pretend to shoot basket, pretend to sing, pretend to sleep, pretend to steer, pretend to steer wheel, pretend to stir, pretend to swing bat, pretend to swipe, pretend to throw, pretend to throw ball, pretend to type, pretend to unlock, pretend to use controller, pretend to wake, pretend to wave, pretend to weigh, pretend to write, pull, push, remote, remove, reveal, right, roll, rollable, rotate, rough, run, scatter, scoop, scroll, separate, shake, shake head, shelf, shout, sip, sit, sleep, slice, slide, smell, smile, smooth, snap, sneeze, speak, spill, spillable, spin, spin dance, spit, spoon, sprinkle, sprint, squeezable, stack, stackable, stand, start, stir, stop, stretch, stumble, surface, swing, swipe, table, tap, taste, tear, throw, tie, tighten, tilt, tomato, top, topple, touch, toy, turn off, turn on, type, uncover, unfasten, unfold, unlock, unplug, unstack, untie, unwrap, upward, vegetable, wake, walk, wall, wave, wet, whisper, window, wrap, write, yawn, zoom in, zoom out \\
\bottomrule
\end{tabular}
\label{tab:concepts}
\end{table}

\newpage
\subsection{Concept set variance}

\begin{wrapfigure}{r}{0.4\textwidth}
    \centering
    \includegraphics[width=\linewidth]{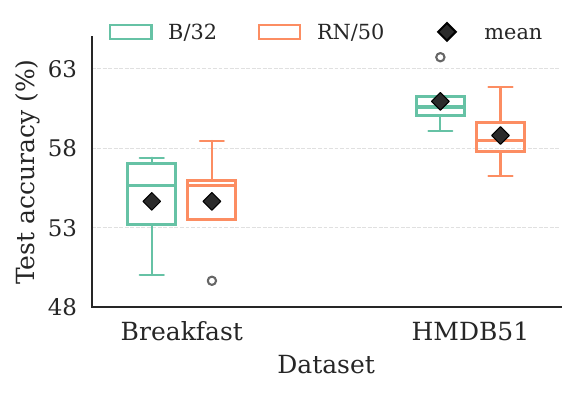}
    \caption{\textbf{Concept set influence.} Distribution of test accuracy across five concept sets. Boxes show the interquartile range, horizontal lines mark medians, diamonds mark means, and the two open circles mark observations beyond the whiskers. }
    \label{fig:concept_set_influence}
\end{wrapfigure}
\textbf{Concept set influence.}
To investigate the influence of the concept proposal set on the performance of MoTIF, we prompt GPT-5 five times, each time requesting a distinct set of candidate concepts (with some natural overlap across runs).
Our experiments demonstrate that given the same prompt to the LLM, concept set affects test accuracy only moderately, with consistent trends across datasets and backbones (Figure~\ref{fig:concept_set_influence}).
In line with prior work on concept bottleneck models, more dataset-specific concepts tend to increase accuracy \citep{discover, dcbm, schrodi}.
However, our main motivation here is not to optimize the quality of the concept proposals, but rather to demonstrate that our framework works broadly and robustly across datasets, backbones, and varying concept sets.
In addition, we note that $C$---the number of concepts retained in the bottleneck---influences the achievable accuracy: very small $C$ restricts expressive power, while very large $C$ may introduce noise and redundancy if concepts are too similar or irrelevant.
Table~\ref{tab:test_concepts} lists the textual concepts used for the Breakfast ablation in Figure~\ref{fig:concept_set_influence}.
We prompted the LLM five times to generate concepts for the CBM using the following prompt:

\begin{verbatim}
Create unique concepts (>100) for a concept-bottleneck model for
the dataset 'Breakfast Actions'. Return them in this format:
"prepare coffee, grind beans, ..."
\end{verbatim}

Both the number and the variety of concepts vary across calls.
Despite this variability, model performance remains stable, demonstrating MoTIF's robustness to different concept sets.
The same prompting procedure was repeated for the other datasets (i.e., HMDB51, UCF101, SSv2).

\subsection{Ablation of concept construction}

To examine how concept vocabulary composition affects performance on SSv2, the dataset in our benchmark that relies most heavily on temporal reasoning, we evaluate five concept sets: (1) nouns only, (2) verbs only, (3) nouns combined with verbs, (4) the union of all concepts, and (5) the original concept set (see Tables~\ref{concept_set_ablation} and~\ref{tab:ablation_concepts}).
Each set is evaluated with MoTIF, MoTIF-ST, and Global CBM. Across all five sets, MoTIF outperforms the matched Global CBM.

\begin{table}[h]
\centering
\caption{\textbf{Concept-set ablation on SSv2.} Top-1 accuracy (\%) with PE-L/14 for four alternative concept sets and the original set.}
\vspace{0.5em}
\begin{tabular}{lccc}
\toprule
\textbf{Concept Set} & \textbf{MoTIF} & \textbf{MoTIF-ST} & \textbf{Global CBM} \\
\midrule
(1) Nouns only          & 18.0 & 20.3 & 14.8 \\
(2) Verbs only          & 24.4 & 28.6 & 21.2 \\
(3) Nouns + Verbs       & 22.1 & 24.5 & 18.2 \\
(4) All concepts        & 29.8 & 36.0 & 26.7 \\
(5) Original set        & 30.0 & 35.1 & 25.5 \\
\bottomrule
\end{tabular}
\label{concept_set_ablation}
\end{table}

The ablation yields three observations:
\begin{itemize}
    \item MoTIF outperforms Global CBM for every concept set, suggesting that temporal modeling improves classification across these vocabularies.
    \item Verb-only concepts (set~2) outperform noun-only concepts (set~1), consistent with the value of action-oriented concepts on SSv2.
    \item The union of all concepts (set~4) gives the highest MoTIF-ST accuracy (36.0\%), while the original set (set~5) gives the highest standard MoTIF accuracy (30.0\%). Thus, the effect of vocabulary composition also depends on the temporal architecture.
\end{itemize}

We also audited class-name similarity in the original, unfiltered global explanations.
At a cosine-similarity threshold of 0.85, the top-ranked concept was similar to a class name in 27.5\%/65.5\% of Breakfast and 76.9\%/90.3\% of HMDB51 explanations for RN/50 and ViT-B/32, respectively. Similarity to the predicted class occurred in 11.3\%/28.2\% and 46.7\%/59.0\% of cases, respectively.
Class-like top explanations are therefore a limitation, particularly on HMDB51.

To test whether the accuracy gains depend on such concepts, we removed every Qwen-generated concept whose text embedding had cosine similarity $\geq 0.85$ to any class name and retrained PE-L/14 models with the original hyperparameters. After filtering, MoTIF remains above Global CBM on Breakfast (+12.8), HMDB51 (+4.0), and SSv2 (+3.2), and is approximately tied on UCF101 ($-$0.6). Thus, the main matched gains largely persist after removing the concepts most similar to class names.

\begin{table}[h]
\centering
\caption{\textbf{MoTIF SSv2 concepts.} Textual concepts used in the concept-set ablation.}
\vspace{0.5em}
\tiny
\begin{tabular}{p{1.4cm} p{0.3cm} p{11cm}}
\toprule
\textbf{Set} & \textbf{Count} & \textbf{Concepts} \\
\midrule
\textbf{Nouns-Only Concepts} & 60 & object, container, box, cup, bowl, plate, spoon, knife, fork, chopstick,
pen, pencil, paper, book, phone, remote, laptop, keyboard, mouse,
bag, backpack, toy, ball, fruit, apple, orange, banana, grape,
vegetable, carrot, cucumber, tomato, bottle, can, lid, cap, key,
lock, door, window, wall, floor, table, chair, shelf, hand, finger,
arm, face, person, background, surface, inside, outside, top, bottom,
left, right, upward, downward\\
\midrule
\textbf{Verbs-Only Concepts}&135 & push, pull, lift, drop, hold, carry, throw, catch, slide, drag,
roll, spin, rotate, flip, fold, unfold, wrap, unwrap, tie, untie,
fasten, unfasten, tighten, loosen, break, cut, slice, chop, tear,
peel, crumple, flatten, bend, stretch, shake, stir, pour, scoop,
sprinkle, stack, unstack, assemble, disassemble, open, close, lock,
unlock, press, tap, swipe, scroll, zoom in, zoom out, point, touch,
wave, clap, knock, snap, swing, juggle, bounce, balance, topple,
insert, remove, fill, empty, mix, separate, spill, scatter, gather,
cover, uncover, hide, reveal, lean, tilt, climb, crawl, jump, hop,
walk, run, sprint, stumble, fall, get up, sit, stand, kneel, crouch,
bow, dance, nod, shake head, smile, frown, laugh, cry, shout,
whisper, speak, yawn, sneeze, cough, sleep, wake, eat, chew, bite,
sip, drink, spit, blow, smell, taste, write, draw, erase, paint,
type, click, drag mouse, plug, unplug, connect, disconnect,
turn on, turn off, start, stop, accelerate, decelerate  \\
\midrule
\textbf{Noun--Verb Concepts} & 104 & lift the box, open the box, close the box, drop the box, push the box, pull the box,
carry the cup, pour from the cup, fill the cup, empty the cup, hold the cup, rotate the cup,
open the bottle, close the bottle, pour from the bottle, lift the bottle, drink from the bottle,
cut the paper, fold the paper, tear the paper, crumple the paper, write on the paper,
open the book, close the book, flip the book, read the book, drop the book,
type on the keyboard, plug in the laptop, unplug the laptop, open the laptop, close the laptop,
click the mouse, drag the mouse, press the key, turn on the phone, turn off the phone,
swipe the phone, tap the phone, scroll on the phone, charge the phone, unlock the phone,
lock the door, unlock the door, open the door, close the door,
lift the chair, move the chair, sit on the chair, stand from the chair,
throw the ball, catch the ball, bounce the ball, roll the ball,
peel the banana, eat the banana, cut the apple, eat the apple, slice the cucumber,
pour the water, stir the soup, mix the ingredients, chop the vegetables,
open the can, close the lid, place the lid, remove the lid,
stack the boxes, unstack the boxes, wrap the gift, unwrap the gift,
tie the rope, untie the rope, press the button, flip the switch, turn the knob,
insert the plug, remove the plug, connect the cable, disconnect the cable,
shake the bottle, squeeze the bottle, pour the juice, stir the drink,
draw on the paper, paint the wall, erase the drawing, fold the towel,
open the backpack, close the backpack, pick up the bag, drop the bag,
open the window, close the window, wipe the table, clean the floor,
open the box with one hand, lift the object, move the object, drop the object,
hold the object, place the object, throw the object, pick up the object\\
\midrule
\textbf{All Concepts} & 391 & push, pull, lift, drop, hold, carry, throw, catch, slide, drag,
roll, spin, rotate, flip, fold, unfold, wrap, unwrap, tie, untie,
fasten, unfasten, tighten, loosen, break, cut, slice, chop, tear,
peel, crumple, flatten, bend, stretch, shake, stir, pour, scoop,
sprinkle, stack, unstack, assemble, disassemble, open, close, lock,
unlock, press, tap, swipe, scroll, zoom in, zoom out, point, touch,
wave, clap, knock, snap, swing, juggle, bounce, balance, topple,
insert, remove, fill, empty, mix, separate, spill, scatter, gather,
cover, uncover, hide, reveal, lean, tilt, climb, crawl, jump, hop,
walk, run, sprint, stumble, fall, get up, sit, stand, kneel, crouch,
bow, dance, spin dance, nod, shake head, smile, frown, laugh, cry,
shout, whisper, speak, yawn, sneeze, cough, sleep, wake, eat, chew,
bite, sip, drink, spit, blow, smell, taste, write, draw, erase,
paint, type, click, drag mouse, plug, unplug, connect, disconnect,
turn on, turn off, start, stop, accelerate, decelerate,
pretend to push, pretend to pull, pretend to pour, pretend to eat, pretend to drink,
pretend to throw, pretend to catch, pretend to type, pretend to swipe, pretend to scroll,
pretend to climb, pretend to fall, pretend to hug, pretend to kiss, pretend to wave,
pretend to play guitar, pretend to drive, pretend to steer, pretend to read, pretend to sleep,
pretend to wake, pretend to write, pretend to draw, pretend to paint, pretend to clean,
pretend to cook, pretend to stir, pretend to measure, pretend to weigh, pretend to look around,
pretend to search, pretend to point, pretend to balance, pretend to open, pretend to close,
pretend to lock, pretend to unlock, pretend to kick, pretend to punch, pretend to block,
pretend to dodge, pretend to jump rope, pretend to row, pretend to paddle, pretend to shoot arrow,
pretend to load gun, pretend to fire gun, pretend to throw ball, pretend to dribble,
pretend to shoot basket, pretend to swing bat, pretend to serve, pretend to catch fish,
pretend to steer wheel, pretend to honk, pretend to use controller, pretend to play piano,
pretend to play drums, pretend to dance, pretend to sing, pretend to clap, pretend to salute,
pretend to bow, pretend to shake hands, pretend to hug, pretend to kiss,
lift the box, open the box, close the box, drop the box, push the box, pull the box,
carry the cup, pour from the cup, fill the cup, empty the cup, hold the cup, rotate the cup,
open the bottle, close the bottle, pour from the bottle, lift the bottle, drink from the bottle,
cut the paper, fold the paper, tear the paper, crumple the paper, write on the paper,
open the book, close the book, flip the book, read the book, drop the book,
type on the keyboard, plug in the laptop, unplug the laptop, open the laptop, close the laptop,
click the mouse, drag the mouse, press the key, turn on the phone, turn off the phone,
swipe the phone, tap the phone, scroll on the phone, charge the phone, unlock the phone,
lock the door, unlock the door, open the door, close the door,
lift the chair, move the chair, sit on the chair, stand from the chair,
throw the ball, catch the ball, bounce the ball, roll the ball,
peel the banana, eat the banana, cut the apple, eat the apple, slice the cucumber,
pour the water, stir the soup, mix the ingredients, chop the vegetables,
open the can, close the lid, place the lid, remove the lid,
stack the boxes, unstack the boxes, wrap the gift, unwrap the gift,
tie the rope, untie the rope, press the button, flip the switch, turn the knob,
insert the plug, remove the plug, connect the cable, disconnect the cable,
shake the bottle, squeeze the bottle, pour the juice, stir the drink,
draw on the paper, paint the wall, erase the drawing, fold the towel,
open the backpack, close the backpack, pick up the bag, drop the bag,
open the window, close the window, wipe the table, clean the floor,
open the box with one hand, lift the object, move the object, drop the object,
hold the object, place the object, throw the object, pick up the object,
object, container, box, cup, bowl, plate, spoon, knife, fork, chopstick,
pen, pencil, paper, book, phone, remote, laptop, keyboard, mouse,
bag, backpack, toy, ball, fruit, apple, orange, banana, grape,
vegetable, carrot, cucumber, tomato, bottle, can, lid, cap, key,
lock, door, window, wall, floor, table, chair, shelf, hand, finger,
arm, face, person, other, background, surface, inside, outside, top,
bottom, left, right, upward, downward, hot, cold, wet, dry, clean,
dirty, empty, full, broken, fixed, smooth, rough, heavy, light,
fragile, durable, rollable, stackable, squeezable, pourable, spillable,
openable, closeable, edible, drinkable  \\
\bottomrule
\end{tabular}
\label{tab:ablation_concepts}
\end{table}

\newpage

\begin{table}[h!]
\centering
\caption{\textbf{MoTIF breakfast concepts.} The textual concepts utilized for the ablation of concept influence.}
\vspace{0.5em}
\tiny
\begin{tabular}{p{0.8cm} p{0.3cm} p{11cm}}
\toprule
\textbf{Set} & \textbf{Count} & \textbf{Concepts} \\
\midrule
Breakfast Original & 223 & add, adjust, apple, arrange, assemble, avocado, bacon, bagel, bake, balance, banana, batter, beat, bin, blend, blender, blow, boil, bottle, bowl, bread, brew, brush, burner, butter, button, carry, carton, catch, cereal, chair, cheese, chop, cinnamon, clap, close, coffee, colander, comb, container, cook, cookbook, cool, core, counter, cover, crack, croissant, cucumber, cup, cupboard, cut, cuttingboard, detergent, dish, drag, drain, drizzle, drop, dry, egg, faucet, fill, flame, flip, fold, fork, freezer, fridge, froth, frown, fruit, fry, garbage, gesture, granola, grate, grater, grill, grind, ham, handle, heat, herb, hide, honey, hood, ice, ingredient, insert, jar, juice, kettle, knife, knob, knock, ladle, laugh, leftover, lid, mash, measure, measuringcup, measuringspoon, milk, mix, mug, napkin, nod, onion, open, orange, oven, ovenmitt, pack, package, pan, pantry, pastry, peel, peeler, pick, pinch, pit, place, plate, plug, point, poke, pour, preheat, press, pull, push, put, reach, recipe, recycle, release, remove, reveal, rinse, roll, rotate, sausage, scale, scoop, scramble, scrub, seal, serve, serving, set, shake, shave, sieve, sink, sip, sit, slice, slide, smile, snap, soap, socket, sort, spatula, spin, sponge, spoon, spread, sprinkle, squeeze, stack, stand, start, steam, steep, stir, stirrer, stool, stop, stove, strawberry, sugar, switch, syrup, table, take, tamp, tap, taste, tea, thermometer, throw, tie, tilt, timer, toast, tomato, tongs, toss, towel, tray, turn, twist, uncover, unfold, unscrew, unstack, untie, unwrap, warm, wash, waste, water, wave, whisk, wipe, wring, yogurt, zest, zip \\
\midrule
Breakfast Set 2&140 & adjust heat, arrange cutlery, bake bread, bake pastry, beat eggs, bend down, bite food, break chocolate, break egg shell, breakfast clock, bubbling liquid, butter toast, chair at table, chew food, chop onion, clean spoon, close carton, close cupboard, close drawer, close fridge, close jar, close microwave, close oven, close tap, cut banana, cut dough, cut sandwich, dice tomato, drip water, drizzle dressing, drizzle honey, drizzle oil, dry dish, dry hands, empty sink, family sitting, flip bread, flip toast, fold mixture, fold napkin, grab fork, grab knife, grab spoon, grate chocolate, grill sandwich, gulp drink, hold bowl, hold glass, hold plate, hold straw, knead dough, lean forward, lick spoon, mash egg, mash potato, melt chocolate, mix salad, mop spill, morning light, move chair, one person eating, open carton, open cupboard, open drawer, open fridge, open microwave, open oven, open tap, peel apple, peel banana, peel egg, peel orange, person standing at stove, place utensil, pour batter, pour carton, preheat oven, press button, put down bowl, put down glass, put down plate, reach cupboard, reach shelf, recycle carton, recycle glass, recycle paper, recycle plastic, rest dough, rinse cup, roast vegetable, roll dough, run water, separate yolk, set napkin, shape dough, shred lettuce, sip drink, sip straw, sit down, sizzling pan, slice cucumber, soap hands, spread batter, spreading butter, spreading jam, spreading topping, sprinkle cheese, sprinkle herbs, sprinkle spices, squeeze lemon, stack plates, stand up from chair, steam rising, stir chocolate, swallow food, take carton, take jar, take package, take utensil, tear package, throw trash, toast bun, toss salad, towel hands, turn knob, turn off kettle, turn off stove, turn on kettle, turn on stove, two people cooking, unwrap sandwich, wash dish, wash hands, water boiling, whisk whites, wipe counter, wipe knife, wipe plate, wrap sandwich, zest lemon \\
\midrule
Breakfast Set 3 & 128 & add cinnamon, add fruit topping, add granola, add honey, add ice to blender, add milk, add milk to cereal, add sugar, arrange cutlery, assemble sandwich, bake pastry, beat eggs, blend smoothie, boil potato, boil water, brew coffee, butter toast, check timer, chop herbs, chop onion, chop vegetables, clear table, close cupboard, close fridge, close jar, close oven, cook bacon, cook pancake, cook sausage, core apple, crack egg, cut sandwich, dice vegetables, drain bacon, drizzle honey, drizzle syrup, dry dishes, fill kettle, flip bacon, flip omelette, flip pancake, follow recipe, froth milk, fry egg, grate cheese, grill sandwich, grind coffee beans, heat pan, insert coffee pod, make omelette, mash avocado, mash potato, measure ingredients, mix batter, open cupboard, open egg carton, open fridge, open jar, open oven, operate espresso machine, pack leftovers, peel banana, peel potato, pick up spoon, pit avocado, place cup, place plate, pour batter, pour cereal into bowl, pour coffee into cup, pour hot water, pour milk, pour smoothie, pour syrup, pour yogurt into bowl, preheat oven, prepare coffee, pull espresso shot, put ingredient in fridge, put leftovers in fridge, read recipe, rinse fruit, scramble eggs, serve pancakes, set table, set timer, sip beverage, slice apple, slice avocado, slice bagel, slice banana, slice bread, slice cheese, slice cucumber, slice ham, slice orange, slice strawberries, slice tomato, spread butter, spread cream cheese, spread jam, spread peanut butter, sprinkle sugar, squeeze lemon, steam milk, steep tea, stir beverage, strain smoothie, take cup, take ingredient from fridge, take leftovers out fridge, take plate, toast bagel, toast bread, unscrew lid, unwrap bread, use fork, use knife, use measuring cup, use measuring spoon, use spatula, use tongs, use whisk, warm croissant, wash dishes, wash fruit, whisk eggs, wipe counter \\
\midrule
Breakfast Set 4& 175 & add cereal, add cinnamon, add cocoa powder, add honey, add ice to blender, add milk, add milk to coffee, add pepper, add salt, add sugar to coffee, add sugar to tea, add toppings, adjust seasoning, arrange cutlery, assemble sandwich, beat eggs, blend smoothie, blow on hot food, boil potato, boil water, brew coffee, butter toast, carry plate to table, check food temperature, check timer, chop herbs, chop onion, chop tomato, clean blender, clean counter, clear table, close cupboard, close egg carton, close fridge, close jar, close microwave, close milk carton, close oven, cook bacon, cook sausage, crack egg, cut lemon, cut sandwich, dice vegetables, drain bacon, drizzle honey, drizzle syrup, dry dishes, dry hands, fill kettle, fill pot with water, flip bacon, flip omelette, flip pancake, follow recipe, froth milk, fry egg, grate cheese, grind coffee beans, hold pan lid, insert bread into toaster, insert coffee pod, make omelette, mash potato, measure ingredients, mix batter, open cupboard, open egg carton, open fridge, open jar, open microwave, open milk carton, open oven, open package, operate blender, operate espresso machine, pack leftovers, peel banana, peel orange, peel potato, pick up cup, pick up knife, pit avocado, place pan off stove, place pan on stove, place plate, pour batter, pour cereal into bowl, pour coffee, pour eggs into pan, pour from carton, pour hot water, pour milk into bowl, pour pancake batter, pour smoothie, pour syrup, pour tea, pour yogurt into bowl, preheat oven, prepare coffee, prepare tea, press coffee, pull espresso shot, put leftovers in fridge, reach for ingredient, read recipe, remove bread from toaster, remove lid from pot, rinse fruit, scoop butter, scramble eggs, search for ingredient, season food, serve omelette, serve pancakes, set table, set timer, sip beverage, sit down, slice apple, slice avocado, slice bagel, slice banana, slice bread, slice cheese, slice cucumber, slice fruit, slice ham, slice kiwi, slice orange, slice pancake stack, slice strawberries, slice tomato, spoon yogurt, spread butter, spread cream cheese, spread jam, spread peanut butter, sprinkle granola, sprinkle sugar, squeeze lemon, stand up, start microwave, steam milk, steep tea, stir coffee, stop microwave, strain smoothie, take leftovers out fridge, take plate, taste food, toast bagel, toast bread, turn off kettle, turn off stove, turn on kettle, turn on stove, unscrew lid, unwrap bread, use fork, use french press, use knife, use measuring cup, use oven mitts, use spatula, use spoon, use toaster, use tongs, use whisk, wash blender, wash dishes, wash fruit, wash hands, whisk eggs, wipe counter \\
\midrule
Breakfast Set 5 & 161 & add cinnamon, add fruit topping, add granola, add honey, add ice to blender, add milk, add milk to cereal, add nuts, add sugar, adjust seasoning, arrange cutlery, assemble sandwich, bake pastry, beat eggs, blend smoothie, blow on hot food, boil potato, boil water, brew coffee, butter toast, carry plate to table, check timer, chop herbs, chop onion, chop vegetables, clear table, close cupboard, close fridge, close jar, close milk carton, close oven, cook bacon, cook pancake, cook sausage, core apple, core pineapple, crack egg, cut parsley, cut pineapple, cut sandwich, dice vegetables, drain bacon, drain can, drizzle honey, drizzle syrup, dry dishes, dry hands, fill kettle, flip bacon, flip omelette, flip pancake, follow recipe, froth milk, fry egg, grate cheese, grill sandwich, grind beans, heat pan, insert coffee pod, juice orange, make omelette, mash avocado, mash potato, measure ingredients, mix batter, open cupboard, open egg carton, open fridge, open jar, open milk carton, open oven, open package, open tin, operate espresso machine, pack leftovers, peel banana, peel orange, peel potato, pick up cup, pick up spoon, pit avocado, place cup, place pan off stove, place plate, pour batter, pour cereal, pour coffee into cup, pour from carton, pour hot water, pour milk, pour smoothie, pour syrup, pour yogurt, preheat oven, prepare coffee, pull espresso shot, put ingredient in fridge, put leftovers in fridge, reach for ingredient, read recipe, remove lid from pot, rinse fruit, scramble eggs, seal container, serve pancakes, set table, set timer, sip beverage, sit down, slice apple, slice avocado, slice bagel, slice banana, slice bread, slice cheese, slice cucumber, slice ham, slice kiwi, slice lemon, slice melon, slice orange, slice pear, slice strawberries, slice tomato, spread butter, spread cream cheese, spread jam, spread peanut butter, sprinkle sugar, squeeze lemon, stand up, steam milk, steep tea, stir beverage, strain smoothie, take cup, take ingredient from fridge, take leftovers out fridge, take plate, taste food, toast bagel, toast bread, toast nuts, unscrew lid, unwrap bread, unwrap package, use fork, use knife, use measuring cup, use measuring spoon, use oven mitts, use spatula, use tongs, use whisk, warm croissant, wash dishes, wash fruit, wash hands, whisk eggs, wipe counter, zest lemon \\
\bottomrule
\end{tabular}
\label{tab:test_concepts}
\end{table}

%% file: sections/checklist.tex
\section*{NeurIPS Paper Checklist}
\begin{enumerate}

\item {\bf Claims}
    \item[] Question: Do the main claims made in the abstract and introduction accurately reflect the paper's contributions and scope?
    \item[] Answer: \answerYes{} 
    \item[] Justification: The claims stated in the abstract are also motivated in the introduction. The remainder of the paper explains MoTIF’s design choices and experimental evaluation, while the appendix provides additional experiments and supporting information.

\item {\bf Limitations}
    \item[] Question: Does the paper discuss the limitations of the work performed by the authors?
    \item[] Answer: \answerYes{} 
    \item[] Justification: 
    We discuss limitations in Section~\ref{discussion} and address the relevant trade-offs throughout the paper.

\item {\bf Theory assumptions and proofs}
    \item[] Question: For each theoretical result, does the paper provide the full set of assumptions and a complete (and correct) proof?
    \item[] Answer: \answerNA{} 
    \item[] Justification: The paper does not present theoretical results that require formal assumptions or proofs.

    \item {\bf Experimental result reproducibility}
    \item[] Question: Does the paper fully disclose all the information needed to reproduce the main experimental results of the paper to the extent that it affects the main claims and/or conclusions of the paper (regardless of whether the code and data are provided or not)?
    \item[] Answer: \answerYes{} 
    \item[] Justification: We list all relevant hyperparameters in Section~\ref{app:hyops} and provide a more detailed description of the architecture through algorithmic overviews.

\item {\bf Open access to data and code}
    \item[] Question: Does the paper provide open access to the data and code, with sufficient instructions to faithfully reproduce the main experimental results, as described in supplemental material?
    \item[] Answer: \answerYes{} 
    \item[] Justification: We provide the full code base for all experiments, from embedding the videos, to training MoTIF and its evaluation.

\item {\bf Experimental setting/details}
    \item[] Question: Does the paper specify all the training and test details (e.g., data splits, hyperparameters, how they were chosen, type of optimizer) necessary to understand the results?
    \item[] Answer: \answerYes{} 
    \item[] Justification: We report all hyperparameter configurations and use the official test splits whenever available. Otherwise, we follow the splits used in prior work and state this explicitly. The data embedding procedure is also documented in the released code.

\item {\bf Experiment statistical significance}
    \item[] Question: Does the paper report error bars suitably and correctly defined or other appropriate information about the statistical significance of the experiments?
    \item[] Answer: \answerYes{} 
    \item[] Justification: We include ablation experiments with multiple random seeds and, where available, evaluate across all train--test splits. We report the resulting variability using the mean and standard deviation, allowing us to assess MoTIF’s robustness across runs.

\item {\bf Experiments compute resources}
    \item[] Question: For each experiment, does the paper provide sufficient information on the computer resources (type of compute workers, memory, time of execution) needed to reproduce the experiments?
    \item[] Answer: \answerYes{} 
    \item[] Justification: We run all experiments on the same hardware and provide an overview of the computational resources and additional runtime statistics in the appendix; see Section~\ref{resources}.
    
\item {\bf Code of ethics}
    \item[] Question: Does the research conducted in the paper conform, in every respect, with the NeurIPS Code of Ethics \url{https://neurips.cc/public/EthicsGuidelines}?
    \item[] Answer: \answerYes{} 
    \item[] Justification: We conform with the NeurIPS Code of Ethics.

\item {\bf Broader impacts}
    \item[] Question: Does the paper discuss both potential positive societal impacts and negative societal impacts of the work performed?
    \item[] Answer: \answerNA{}
    \item[] Justification: The paper primarily focuses on the methodological and empirical aspects of MoTIF. While the motivation of concept-based modeling is connected to positive societal impact through improved inspectability and controllability, we do not provide a dedicated discussion of positive societal impacts.
    
\item {\bf Safeguards}
    \item[] Question: Does the paper describe safeguards that have been put in place for responsible release of data or models that have a high risk for misuse (e.g., pre-trained language models, image generators, or scraped datasets)?
    \item[] Answer: \answerNA{} 
    \item[] Justification: MoTIF and its code does not contain such risks.

\item {\bf Licenses for existing assets}
    \item[] Question: Are the creators or original owners of assets (e.g., code, data, models), used in the paper, properly credited and are the license and terms of use explicitly mentioned and properly respected?
    \item[] Answer: \answerYes{} 
    \item[] Justification: We provide proper attribution for all existing assets used in the paper, including datasets, pretrained models, and relevant codebases. The code and MoTIF models introduced by this work are created by the authors. We also document and respect the applicable licenses and terms of use for the datasets used for scientific research.

\item {\bf New assets}
    \item[] Question: Are new assets introduced in the paper well documented and is the documentation provided alongside the assets?
    \item[] Answer: \answerYes{} 
    \item[] Justification: We provide next to the paper also the code which as a documentation for reproducing the results. 

\item {\bf Crowdsourcing and research with human subjects}
    \item[] Question: For crowdsourcing experiments and research with human subjects, does the paper include the full text of instructions given to participants and screenshots, if applicable, as well as details about compensation (if any)? 
    \item[] Answer: \answerYes{} 
    \item[] Justification: Section~\ref{sec:explanations} reports this supplementary eight-rater preference study. The supplementary material provides the full instructions given to raters, and the rating interface. Raters were asked only to choose which explanation they found more useful, or to indicate a tie; no personal information was collected from them. All raters were compensated at a rate above the minimum wage in the country where the study was conducted (Germany).
    

\item {\bf Institutional review board (IRB) approvals or equivalent for research with human subjects}
    \item[] Question: Does the paper describe potential risks incurred by study participants, whether such risks were disclosed to the subjects, and whether Institutional Review Board (IRB) approvals (or an equivalent approval/review based on the requirements of your country or institution) were obtained?
    \item[] Answer: \answerYes{} 
    \item[] Justification: The eight-rater preference study posed minimal foreseeable risk beyond participants’ time commitment. Participants were informed about the task and its voluntary nature, and no personal information was collected.

\item {\bf Declaration of LLM usage}
    \item[] Question: Does the paper describe the usage of LLMs if it is an important, original, or non-standard component of the core methods in this research? Note that if the LLM is used only for writing, editing, or formatting purposes and does \emph{not} impact the core methodology, scientific rigor, or originality of the research, declaration is not required.
    \item[] Answer: \answerYes{} 
    \item[] Justification: We used established coding agents such as Codex and Claude Code to support implementation. However, no LLM-based tool was used to generate the scientific contribution of the paper.

\end{enumerate}